\documentclass[preprint,5p,times,twocolumn,fleqn]{elsarticle}
\usepackage{amsmath,amssymb}
\usepackage{tabularx,array}
\usepackage{enumitem}
\usepackage{makecell}
\usepackage{needspace}
\usepackage{caption}
\usepackage[ruled,vlined]{algorithm2e}
\usepackage{booktabs,multirow}
\usepackage[hidelinks]{hyperref}

\usepackage{float}
\usepackage{verbatim} %comments
\usepackage{stfloats}
\usepackage{placeins}

\newcolumntype{Y}{>{\centering\arraybackslash}X} % centered X (tabularx)

\biboptions{numbers,sort&compress}

\begin{document}
\begin{frontmatter}

%The title page must contain the title of the paper and the full name/full affiliation with country/e-mail address for each author and co-author of the manuscript. Please make sure you have included all elements listed below with your manuscript submission.

\title{CORE-Acu: Structured Reasoning Traces and Knowledge Graph Safety Verification for Acupuncture Clinical Decision Support}

% 作者信息
\author[aff1]{Liuyi Xu}
\ead{xuliuyi@mails.neu.edu.cn}

\author[aff1]{Yun Guo}
\ead{guoyun@mails.neu.edu.cn}

\author[aff1]{Ming Chen}
\ead{chenm4@mails.neu.edu.cn}

\author[aff6]{Zihan Dun}
\ead{dunzihan@stumail.ysu.edu.cn}

\author[aff5]{Yining Qian\corref{cor1}}
\ead{qianyiningning@126.com}

\author[aff1]{An-Yang Lu}
\ead{luanyang@mail.neu.edu.cn}

\author[aff2]{Shuang Li}
\ead{shuangliai@buaa.edu.cn}

\author[aff3,aff4]{Lijun Liu}
\ead{liulijun@mail.neu.edu.cn}

\cortext[cor1]{Corresponding author}

\address[aff1]{College of Information Science and Engineering, Northeastern University, Shenyang 110819, China}

\address[aff6]{School of Information Science and Engineering, Yanshan University, Qinhuangdao, Hebei 066000, China}

\address[aff5]{School of Computer Science and Engineering, Northeastern University, Shenyang 110819, China}

\address[aff2]{School of Artificial Intelligence, Beihang University, Beijing 100000, China}

\address[aff3]{Key Laboratory of Bioresource Research and Development of Liaoning Province, Northeastern University, Shenyang 110169, China}

\address[aff4]{College of Life and Health Sciences, Northeastern University, Shenyang 110169, China}

\begin{abstract}
Large language models (LLMs) show significant potential for clinical decision support (CDS), yet their black-box nature---characterized by untraceable reasoning and probabilistic hallucinations---poses severe challenges in acupuncture, a field demanding rigorous interpretability and safety. 
To address this, we propose CORE-Acu, a neuro-symbolic framework for acupuncture clinical decision support that integrates Structured Chain-of-Thought (S-CoT) with knowledge graph (KG) safety verification.
First, we construct the first acupuncture Structured Reasoning Trace dataset and a schema-constrained fine-tuning framework. By enforcing an explicit causal chain from pattern identification to treatment principles, treatment plans, and acupoint selection, we transform implicit Traditional Chinese Medicine (TCM) reasoning into interpretable generation constraints, mitigating the opacity of LLM-based CDS. 
Furthermore, we construct a TCM safety knowledge graph and establish a ``Generate--Verify--Revise'' closed-loop inference system based on a Symbolic Veto Mechanism, employing deterministic rules to intercept hallucinations and enforce hard safety boundaries.
Finally, we introduce the Lexicon-Matched Entity-Reweighted Loss (LMERL), which corrects terminology drift caused by the frequency--importance mismatch in general optimization by adaptively amplifying gradient contributions of high-risk entities during fine-tuning.
Experiments on 1,000 held-out cases demonstrate CORE-Acu's superior entity fidelity and reasoning quality. Crucially, CORE-Acu achieved 0/1,000 observed safety violations (95\% CI: 0--0.37\%), whereas GPT-4o exhibited an 8.5\% violation rate under identical rules. 
These results establish CORE-Acu as a robust neuro-symbolic framework for acupuncture clinical decision support, guaranteeing both reasoning auditability and strict safety compliance.
\end{abstract}


% \nonumnote{This work was supported in part by the National Natural Science Foundation of China (NSFC) under Grant Nos. 62522309, 31871363, and 31900575, and in part by the Natural Science Foundation of Liaoning Province under Grant No. 2025JH6/101000012.}
\end{frontmatter}

\section{Introduction}
\label{introduction}
In medical natural language processing (NLP), large language models (LLMs) have progressed from general-purpose question answering to supporting complex clinical workflows, including clinical text summarization, evidence retrieval, and clinical decision support (CDS)~\citep{li_ehr_scoping_2024,jiang_health_2023}. 
Despite strong performance of domain-adapted models on standardized benchmarks~\citep{azizi_medpalm2_2025,nori_gpt4_medical_2023}, their deployment in high-stakes CDS faces a fundamental trust gap. 
This gap arises from an inherent mismatch: LLMs are primarily optimized for probabilistic next-token prediction, which offers neither determinism nor transparent, auditable reasoning. 
While probabilistic fluency may be adequate for open-domain dialogue, clinical settings require outputs with internally consistent reasoning and safety guarantees~\citep{mesko_regulatory_2023}. 
Without explicit logical constraints, even strong medical LLMs can produce hallucinations---linguistically plausible yet clinically incorrect recommendations---thereby jeopardizing patient safety~\citep{goh_llm_diagnostic_reasoning_2024,pandit_medhallu_2025}.

These challenges are further amplified in Traditional Chinese Medicine (TCM). 
Unlike modern medicine, which is guided by standardized clinical guidelines, TCM clinical practice hinges on a highly contextualized reasoning process known as syndrome differentiation and treatment (\textit{bianzheng lunzhi}). 
Bridging the gap between symptoms and prescriptions typically requires critical intermediate steps, including syndrome identification (Diagnosis), pathogenesis analysis (Pathology), and the formulation of therapeutic principles (Principle). 
However, existing large models often bypass these implicit reasoning steps and instead attempt to learn a direct mapping from symptoms to prescriptions. 
This black-box generation paradigm not only makes the reasoning process non-auditable but also frequently yields reasoning--output discrepancies---cases in which an apparently appropriate prescription is produced despite flawed intermediate reasoning---thereby undermining clinical usability~\citep{huang_tcm-3ceval_2025}.

Furthermore, among TCM subdomains, acupuncture imposes unique challenges regarding contraindications due to its nature as an invasive physical intervention. 
Unlike herbal prescriptions, acupuncture entails physical manipulation at specific anatomical locations, requiring strict adherence to exclusion criteria to prevent medical malpractice.
Failures to respect these ``red lines'' can directly endanger patient safety. 
For instance, in pregnancy-related cases, prescribing Hegu (LI4)---an acupoint traditionally contraindicated due to its association with promoting uterine activity---may increase the risk of severe adverse outcomes. 
Despite these high stakes, standard LLMs relying on unconstrained probabilistic decoding lack robust mechanisms to enforce non-negotiable clinical constraints.
Consequently, generated prescriptions remain vulnerable to uncontrolled safety hazards, as neural generation alone provides limited guarantees for strictly suppressing rule-violating outputs.

Beyond strict prohibitions, the high-fidelity handling of safety-critical entities---including acupoint identifiers and meridian attribution---presents a precision challenge.
In this context, minor terminological discrepancies can lead to disproportionately severe consequences. 
For example, distinguishing Taixi (KI3) from Taiyuan (LU9) hinges on a single character, yet their therapeutic indications and meridian affiliations differ substantially. 
However, standard optimization objectives in general LLMs, such as cross-entropy loss, typically assign uniform weighting to all tokens, creating a frequency--importance mismatch.
The training signal penalizes mispredicting a safety-critical acupoint name no more than mispredicting a common stopword. 
This mismatch exacerbates terminology drift and weakens the model's ability to precisely generate high-risk entities amidst common vocabulary.

To address these gaps, we propose CORE-Acu (Constrained Ontology Reasoning Engine for Acupuncture), a neuro-symbolic framework that integrates structured reasoning traces with knowledge graph (KG) safety verification. 
Specifically, we construct structured reasoning traces to illuminate the reasoning black box, deploy a symbolic veto mechanism to intercept safety violations, and introduce re-weighted learning to rectify entity imprecision.
Our contributions are threefold:

\begin{itemize}
    \item \textbf{Structured Chain-of-Thought (S-CoT) and Schema-Constrained Fine-Tuning:} 
    We construct the first acupuncture S-CoT dataset and introduce a schema-constrained fine-tuning paradigm. 
    By enforcing an explicit causal chain from pattern identification to therapeutic principles and acupoint selection, we convert implicit TCM reasoning into interpretable, schema-compliant outputs, thereby mitigating the black-box limitation.

    \item \textbf{Symbolic Veto Mechanism for Hard Safety Boundaries:}
    We propose a neuro-symbolic governance module featuring a Symbolic Veto Mechanism.
    By constructing a dedicated TCM contraindication KG as the symbolic substrate, the module orchestrates a \textit{Generate--Verify--Revise} inference loop.
    This mechanism leverages deterministic rules to intercept rule-violating generations (e.g., contraindicated acupoints) in real time, strictly enforcing hard safety constraints.

    \item \textbf{Lexicon-Matched Entity-Reweighted Loss (LMERL):} 
    We propose LMERL to address the terminology drift induced by the frequency-importance mismatch. 
    By injecting domain priors during fine-tuning, this mechanism adaptively up-weights gradients for high-risk entities, thereby improving the precision and robustness of safety-critical terminology generation.
\end{itemize}

\section{Related Work}
\subsection{Large Language Models in TCM and Clinical Constraints}

Although LLMs have demonstrated strong capabilities in clinical interaction and clinical text summarization~\citep{singhal_medpalm_2023}, their use in high-stakes clinical decision support remains limited by hallucinations and poor reasoning traceability. 
Within TCM, recent efforts have developed domain-adapted models---including HuatuoGPT~\citep{zhang_huatuogpt_2023}, Zhongjing~\citep{yang_zhongjing_2023}, and BianCang~\citep{wei_biancang_2025}---primarily via instruction fine-tuning on large-scale medical dialogues and classical TCM texts. 
However, most existing work centers on herbal prescription generation or general patient inquiry, leaving safety-critical procedural subdomains comparatively underexamined.

In particular, acupuncture diagnosis and prescription remain underexplored and present distinct challenges~\citep{cheng_tcm-eval_2025,huang_tcm-3ceval_2025}. 
Unlike herbal medicine, acupuncture involves invasive, needle-based interventions at specific anatomical locations. 
This setting requires a rigorous and interpretable reasoning chain that explicitly links syndrome differentiation, therapeutic principles, and precise acupoint selection. 
Yet, existing approaches often collapse this multi-step process into a direct symptom-to-acupoint mapping~\citep{li_acukg_2025}. 
Without transparent diagnostic logic and auditable intermediate reasoning, such black-box models are unlikely to satisfy the safety requirements for clinical adoption or earn physician trust.

To reduce generation risks, recent studies have explored integrating retrieval-augmented generation (RAG) with KG~\citep{he_opentcm_2025}. 
While these methods can improve evidence grounding, KGs are typically used as retrieval resources rather than as symbolic constraints for safety verification. 
In high-stakes settings such as acupuncture, probabilistic grounding alone is insufficient; explicit, rule-based governance mechanisms are needed to enforce contraindication rules (e.g., pregnancy-related restrictions). 
This gap motivates our proposed neuro-symbolic framework.

\subsection{Optimization Objectives in Domain Adaptation}

Although parameter-efficient fine-tuning (PEFT) has become a de facto standard for adapting LLMs~\citep{houlsby_adapters_2019,hu_lora_2022,dettmers_qlora_2023}, most approaches retain the same cross-entropy objective inherited from general pre-training. 
This choice induces a \textit{frequency--importance mismatch} in clinical generation: under uniform token weighting (by default), the training signal encourages models to fit high-frequency function words while under-penalizing errors on sparse yet safety-critical entities~\citep{lin_focal_loss_2017,cui_class_balanced_2019}. 
Although keyword-aware objectives have been explored in speech recognition and information retrieval~\citep{pineiro_weighted_ce_asr_2024,gandhe_optimization_2014,kwok_improving_2025}, entity-driven gradient reweighting remains under-studied for long-context clinical reasoning, particularly when strict terminological precision is required in acupuncture~\citep{zhang_show_2023}.

\subsection{Evaluation}
Reliable evaluation of accuracy and safety for generation in high-risk medical settings remains challenging. Traditional $n$-gram metrics such as BLEU~\citep{papineni_bleu_2002} and ROUGE~\citep{in_rouge_2004} primarily measure surface lexical overlap and are poorly aligned with clinical logic or semantic consistency~\citep{li_chatdoctor_2023}. As a result, unsafe or logically flawed recommendations can still obtain high overlap scores, limiting their utility for safety auditing.

Recently, LLM-as-a-judge has been used to score generation quality with strong foundation models~\citep{li_chatdoctor_2023,zheng2023mtbench,liu_geval_2023}. However, generic judges may lack domain-specific nuances in TCM, and can mis-evaluate complex acupuncture reasoning without expert calibration.

RAG and knowledge-grounding can improve input evidence, but do not guarantee faithful use of retrieved information~\citep{wang_knowledge-tuning_2025}. Studies on RAG faithfulness report knowledge conflicts, where models prefer erroneous parametric memory over provided evidence~\citep{shi_trusting_2023,liu_lost_middle_2023}; models may also underutilize retrieved context due to position effects or limited reasoning, leading to unfaithful hallucinations despite accurate inputs. These findings suggest that input-side enhancement alone is insufficient for clinical safety, motivating post-hoc verification.

Moreover, prevailing evaluation pipelines predominantly terminate at the scoring phase, failing to offer actionable mechanisms for error rectification. While recent studies have explored intrinsic self-correction strategies~\citep{madaan_self_refine_2023,gao_rarr_2023}, relying solely on a model's latent behavior often yields unstable results. Consequently, there remains a lack of robust frameworks capable of translating evaluation feedback into deterministic revision instructions. Accordingly, we introduce a verifiable \textit{Generate--Verify--Revise} closed-loop framework. By integrating multi-source adjudication with explicit rewriting protocols, our approach is designed to not only identify safety risks but also facilitate their systematic mitigation.

\section{Methodology}
\label{sec:methodology}

\begin{figure*}[t] 
    \centering
    \includegraphics[width=\textwidth]{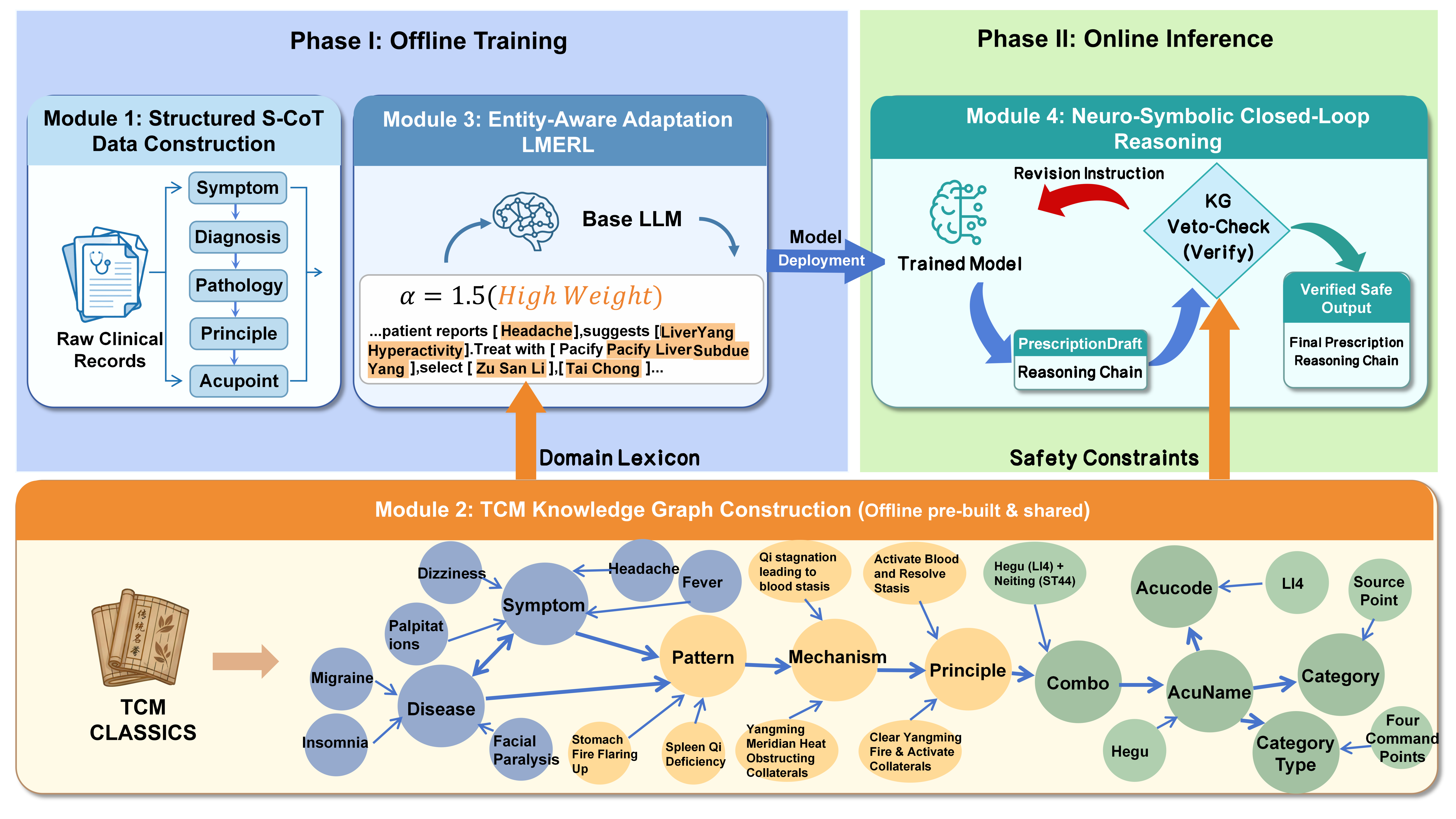} 

    \caption{\textbf{Overall architecture of the CORE-Acu framework.} The framework operates in two phases: (1) \textbf{Phase I (Offline Training):} \textbf{Module 1} constructs the S-CoT corpus to model the clinical causal chain, while \textbf{Module 2} builds the TCM KG. These inputs guide \textbf{Module 3} to fine-tune the LLM via LMERL, focusing on safety-critical entities. (2) \textbf{Phase II (Online Inference):} \textbf{Module 4} implements a neuro-symbolic closed-loop mechanism, leveraging KG constraints from Module 2 to verify prescriptions and trigger bounded self-correction for guaranteed safety.}
    \label{fig:framework}
\end{figure*}

The overall framework of CORE-Acu, illustrated in Fig.~\ref{fig:framework}, integrates structured reasoning traces with knowledge-graph-based safety verification. 
The framework operates through a \textit{Construction--Adaptation--Verification} lifecycle, designed to approach the limitations of unstructured black-box models identified in the introduction.

The architecture consists of four tightly coupled components designed to operationalize clinical reasoning logic:

\begin{enumerate}
    \item \textbf{Structured Reasoning Trace Construction (Sec.~\ref{sec:scot_dataset}):} We first address the lack of traceability by constructing a S-CoT dataset (Module 1). This component explicitly maps the causal pathway from symptoms to acupoints, transforming implicit reasoning into observable logic.
    \item \textbf{TCM Knowledge Graph Construction (Sec.~\ref{sec:kg}):} To provide a rigid safety boundary, we build a domain-specific KG (Module 2). This symbolic substrate serves as the ground truth for acupoint attributes and contraindication rules, enabling deterministic verification.
    \item \textbf{Entity-Aware Adaptation (Sec.~\ref{sec:lmerl}):} To mitigate entity-level hallucinations, we introduce the LMERL (Module 3). This mechanism aligns the model's optimization focus with high-stakes clinical terminology during instruction tuning.
    \item \textbf{Neuro-Symbolic Closed-Loop Reasoning (Sec.~\ref{sec:kg_infer}):} 
    Finally, we develop a neuro-symbolic closed-loop system (Module 4) to operationalize the safety governance. 
    Unlike static filtering, this module implements a dynamic \textit{Generate--Verify--Revise}  reasoning cycle. 
    By leveraging the TCM-KG as a symbolic supervisor, the system iteratively detects potential contraindication violations and provides feedback to the neural generator, thereby striving to achieve a self-corrected and safety-compliant output.
\end{enumerate}

By integrating these components, CORE-Acu transforms the generation process from an open-ended probability distribution into a bounded, verifiable clinical decision workflow.

% \begin{figure*}[t] 
%     \centering
%     % \includegraphics[width=\textwidth]{figure1_architecture.pdf} 
%     \includegraphics[width=\textwidth]{figure1.pdf} 
%     \caption{Overall architecture of the CORE-Acu framework.}
%     \label{fig:architecture}
% \end{figure*}

% \begin{figure*}[t]
%     \centering
%     \includegraphics[width=\textwidth]{figure2.pdf} 
%     \caption{The Triangulated Neuro-Symbolic Validation Framework of CORE-Acu.}
%     \label{fig:Validation}
% \end{figure*}
\begin{figure*}[t]
    \centering
    \includegraphics[width=\textwidth]{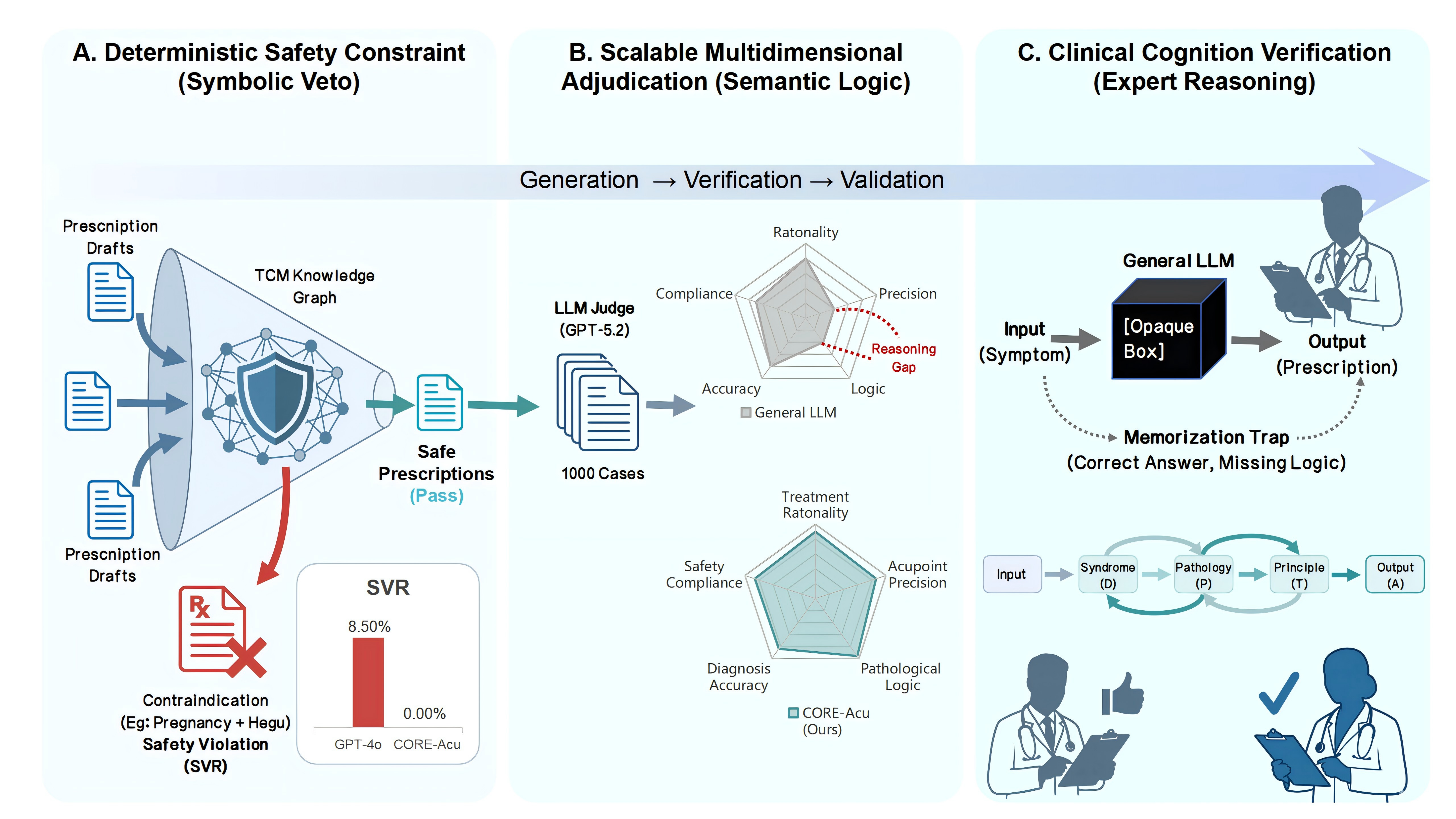} 
    \caption{\textbf{The Triangulated Neuro-Symbolic Validation Framework of CORE-Acu.} The protocol begins with a Symbolic Veto Mechanism that utilizes the TCM KG to deterministically intercept hard safety violations, effectively reducing the Safety Violation Rate. Validated outputs then undergo multidimensional adjudication and expert cognitive verification, which scrutinize the causal reasoning chain to distinguish authentic clinical logic from black-box memorization traps.}

    \label{fig:Validation}
\end{figure*}

\subsection{S-CoT Dataset Construction and Schema}
\label{sec:scot_dataset}

To support verifiable and interpretable acupuncture clinical reasoning, we constructed Acu-Reasoning, which is, to our knowledge, the first large-scale acupuncture S-CoT dataset comprising 42,512 samples. Unlike traditional medical corpora that rely on shallow instruction--response pairs, this dataset explicitly encodes the complete causal chain from symptom reporting to acupoint selection (see Table~\ref{tab:example_instance} for a detailed case instance).

\subsubsection{Causal Modeling of Acupuncture Prescription}
We model the acupuncture prescription task as a sequential decision process rather than a direct mapping. Traditional Supervised Fine-Tuning (SFT) typically optimizes a conditional probability $P(A|C)$, directly mapping clinical complaints $C$ to prescriptions $A$. However, this shallow mapping encourages models to learn statistical shortcuts, bypassing the critical diagnostic logic required for safety.

To address this, we designed a strict schema to enforce the learning of latent reasoning states: \textit{Bian Zheng} (Diagnosis, $D$) $\rightarrow$ \textit{Bing Ji} (Pathology, $P$) $\rightarrow$ \textit{Zhi Ze} (Principle, $T$) $\rightarrow$ \textit{Xuan Xue} (Acupoints, $A$). By explicitly modeling these intermediate variables, we factorize the generation probability according to the clinical causal order:

\begin{equation}
\label{eq:scot_factor}
\begin{split}
    & P_\Theta(D,P,T,A \mid C) \\
    & = \underbrace{P(D|C)P(P|C,D)}_{\text{Bian Zheng (Diagnosis)}} \cdot \underbrace{P(T|C,D,P)}_{\text{Lun Zhi (Planning)}} \cdot \underbrace{P(A|D,P,T)}_{\text{Xuan Xue (Selection)}}
\end{split}
\end{equation}

where $\Theta$ represents the model parameters. This factorization imposes a structural constraint: the model cannot generate acupoints $A$ based solely on symptoms $C$. Instead, it must first resolve the intermediate variables $(D, P, T)$, thereby making the reasoning process observable and machine-verifiable. The detailed schema definition is provided in Table~\ref{tab:schema}.

\subsubsection{Data Curation Pipeline}
To construct this high-quality corpus in alignment with the proposed causal model, we implemented a rigorous three-stage human-in-the-loop pipeline:

\begin{enumerate}
    \item \textbf{Heterogeneous Source Aggregation:} We aggregated raw clinical records from authoritative TCM case report collections, textbooks, and classical literature. These sources are highly unstructured, often containing implicit reasoning.
    \item \textbf{Logic-Driven Restructuring:} This was the most labor-intensive phase. Since raw records often omit intermediate steps obvious to experts, we employed a semi-automated inference-and-verify strategy. We utilized a foundation model (GPT-5.2) prompted with our strict clinical schema to extract explicit fields and, crucially, to \textit{infer} missing logical links (e.g., deducing the therapeutic principle implied by a specific diagnosis). This transformed implicit clinical narratives into fully explicit reasoning chains.
    \item \textbf{Expert Causal Verification:} A panel of three senior TCM practitioners reviewed a stratified sample (10\%) of the dataset. Unlike standard QA checks, this review focused on causal consistency: verifying that the \texttt{diagnosis} logically supported the \texttt{principle}, and that the \texttt{acupoints} were valid implementations of that principle. Errors in entity extraction or causal logic were corrected, and these corrections served as feedback to iteratively refine the extraction prompts.
\end{enumerate}

The resulting dataset serves as the backbone for supervising the model's reasoning capabilities.

\subsection{TCM Knowledge Graph Construction for Symbolic Safety Governance}
\label{sec:kg}

To complement the probabilistic reasoning of S-CoT with a \textit{deterministic substrate} for intercepting hallucinations, we constructed a domain-specific KG, which serves as the symbolic ground truth for a Veto Mechanism. Unlike traditional TCM knowledge bases designed for general information retrieval, our schema prioritizes constraint modeling, aiming to translate hard safety boundaries into computable logic.

\subsubsection{Ontology Standardization and Lexical Anchoring}
To establish an authoritative auditing standard, we strictly aligned the graph schema with the \textit{WHO International Standard Terminologies} and the National Standard \textit{GB/T 12346-2006}. This rigorous standardization serves dual critical functions: 
(1) It resolves terminological ambiguity (e.g., synonymy risks), preventing the model from bypassing safety checks due to naming variations; 
(2) It provides Lexical Anchors for our Lexicon-Matched Entity-Reweighted Loss (LMERL, see Sec.~\ref{sec:lmerl}), ensuring consistency between the training objective and the verification rules.

\subsubsection{Expert-Curated Constraint Modeling}
A key innovation in our pipeline is the transformation of unstructured clinical warnings into machine-verifiable graph relations. Through Expert-Curated Constraint Modeling, we explicitly encoded over 1,200 hard constraints extracted from classical texts and modern guidelines. These are represented not merely as text attributes, but as directed logical edges to enforce a symbolic veto:

\begin{itemize}
    \item \textbf{Conditional Prohibition Edges:} These encode dynamic constraints where an acupoint is generally safe but strictly prohibited under specific patient states (e.g., $LI4 \xrightarrow{\texttt{ProhibitedFor}} \textit{Pregnancy}$).
    \item \textbf{Anatomical Risk Attributes:} Nodes are enriched with physical metadata (e.g., \texttt{DeepNeedlingRisk}, \texttt{VitalOrganProximity}) to flag high-risk operations requiring secondary confirmation.
\end{itemize}

Structurally, the KG is organized into three functional layers—Diagnostic, Therapeutic, and Physical Safety—to mirror the S-CoT reasoning chain, enabling step-by-step verification from diagnosis to intervention. The final graph comprises 4,628 nodes and 12,500 edges (Table~\ref{tab:kg_schema}), covering approximately 90\% of routine clinical pathways.

\begin{table*}[t] 
    \centering
    \caption{\textbf{Schema Definition and Statistics of the Proposed TCM-KG.} The graph integrates standardized acupoint attributes with expert-verified clinical reasoning logic.}
    \label{tab:kg_schema}
    
    \normalsize 
    \renewcommand{\arraystretch}{1.2}
    
    \begin{tabularx}{\textwidth}{l l X c}
        \toprule
        \textbf{Category} & \textbf{Node Entity} & \textbf{Semantics Description} & \textbf{Count} \\
        \midrule
        
        \multirow{4}{*}{\textbf{Basic Info}} 
          & AcuCode & Standard alphanumeric code (e.g., LI4). & 361 \\
          & AcuName & Standard Romanized/Pinyin name. & 361 \\
          & Category & The Meridian or Channel attribution. & 14 \\
          & CategoryType & Functional classification. & 28 \\
        \midrule
        
        \multirow{3}{*}{\textbf{Clinical}} 
          & Disease & Western or Traditional disease names. & 520 \\
          & Symptom & Discrete clinical symptoms or complaints. & 1,240 \\
          & Pattern & TCM Syndromes (Bian-Zheng / Zheng). & 192 \\
        \midrule
        
        \multirow{2}{*}{\textbf{Reasoning}} 
          & Mechanism & Pathological mechanism explaining illness. & 850 \\
          & Principle & Therapeutic principles guiding treatment. & 410 \\
        \midrule
        
        \textbf{Treatment} 
          & Combo & Validated acupoint combinations. & 652 \\
        \midrule
        
        \multirow{2}{*}{\textbf{Total}} 
          & \multicolumn{2}{l}{\textbf{Total Nodes (Entities)}} & \textbf{4,628} \\
          & \multicolumn{2}{l}{\textbf{Total Relations (Edges)}} & \textbf{12,500} \\
        \bottomrule
    \end{tabularx}
\end{table*}

\subsection{Domain-Adaptive Optimization via Lexicon-Matched Entity-Reweighted Loss}
\label{sec:lmerl}

To address the fundamental Frequency--Importance Mismatch where the information density of acupuncture is inversely distributed relative to the natural long-tail token frequency (Zipf's law)—we propose the LMERL. This objective functions as a domain-adaptive filter, explicitly reshaping the gradient landscape to focus on rare yet safety-critical entities that are typically overwhelmed by high-frequency functional words. Unlike standard Cross-Entropy (CE) objectives that treat all tokens uniformly and prioritize linguistic fluency, LMERL aligns the optimization focus with clinical importance rather than statistical frequency.

\subsubsection{Mathematical Formulation}
Let $\mathcal{D} = \{(\mathbf{x}, \mathbf{y})\}$ denote the training corpus, where $\mathbf{y} = (y_1, \dots, y_T)$ is the structured S-CoT response. We utilize the domain lexicon $\mathcal{V}_{\text{dom}}$ (derived from the KG in Sec.~\ref{sec:kg}) as the set of \textit{Lexical Anchors}.

To decouple optimization priority from token frequency, LMERL assigns a dynamic importance score $\omega(y_t)$ to each token. The optimization objective is formalized as the expected entity-weighted negative log-likelihood:

\begin{equation}
\label{eq:lmerl}
\mathcal{L}_{\text{LMERL}}(\Theta) = - \mathbb{E}_{(\mathbf{x}, \mathbf{y}) \sim \mathcal{D}} \left[ \frac{1}{Z(\mathbf{y})} \sum_{t=1}^{T} \omega(y_t) \cdot \log P_{\Theta}(y_t \mid \mathbf{x}, \mathbf{y}_{<t}) \right]
\end{equation}
where $Z(\mathbf{y}) = \sum_{t=1}^{T} \omega(y_t)$ serves as a sequence-level normalization factor to prevent gradient scale drift across samples with varying entity densities.

Crucially, the importance function $\omega(y_t)$ is designed as a piecewise step function to selectively amplify the gradient signal for long-tail domain terms:
\begin{equation}
\label{eq:weight_func}
\omega(y_t) = 
\begin{cases} 
\alpha & \text{if } y_t \in \mathcal{V}_{\text{dom}} \quad (\text{Domain Anchors}) \\
1 & \text{otherwise} \quad (\text{General Syntax})
\end{cases}
\end{equation}
where $\alpha > 1$ is a hyperparameter governing the intensity of the domain bias.

\subsubsection{Gradient Focusing in Parameter-Efficient Tuning}
This reweighting strategy provides a necessary inductive bias for Low-Rank Adaptation (LoRA). Since LoRA updates are constrained to low-rank matrices $\Delta\Theta$ with a limited parameter budget, standard uniform optimization often results in these scarce parameters being diluted by general syntactic fitting. 
By setting $\alpha=1.5$ (determined via sensitivity analysis, see Sec.~\ref{sec:sensitivity_alpha}), LMERL effectively implements Gradient Focusing: it steers the updates of $\Delta\Theta$ specifically toward the manifold of domain-critical entities, ensuring that the adaptation capacity is allocated to establishing precise clinical mappings rather than general linguistic redundancy.

\subsection{Neuro-Symbolic Closed-Loop Governance}
\label{sec:kg_infer}

Standard LLM inference acts as an open-loop process: once a token is generated, it is typically immutable. In high-stakes medicine, this ``generate-and-hope'' paradigm is unacceptable. We therefore implement a Closed-Loop Neuro-Symbolic Control System that treats the LLM's output not as a final decision, but as a tentative proposal subject to rigorous symbolic auditing.

\subsubsection{Symbolic Constraint Definition}
We define the safety boundary using the KG constructed in Sec.~\ref{sec:kg}. Let $\mathcal{G}=(\mathcal{V}, \mathcal{E})$ be the KG. We isolate a subset of edge types $\mathcal{R}_{\text{unsafe}} \subset \mathcal{E}$ representing clinically prohibitive relations, such as point-to-point antagonisms (\texttt{IncompatibleWith}) and state-dependent contraindications (\texttt{ProhibitedForCondition}).

Formally, given a patient context $\mathcal{C}$ (extracted conditions) and a candidate prescription $\mathcal{A}$ (generated acupoints), the Violation Set $V$ is computed via graph intersection operations:
\begin{equation}
\label{eq:violation_set}
V(\mathcal{C}, \mathcal{A}) = \underbrace{\left[ (\mathcal{A} \times \mathcal{R}_{\text{unsafe}} \times \mathcal{A}) \cap \mathcal{E} \right]}_{\text{Internal Conflicts}} \;\cup\; \underbrace{\left[ (\mathcal{C} \times \mathcal{R}_{\text{unsafe}} \times \mathcal{A}) \cap \mathcal{E} \right]}_{\text{Contraindications}}
\end{equation}

The proposal is deemed unsafe if $V(\mathcal{C}, \mathcal{A}) \neq \emptyset$.

\subsubsection{The Generate--Verify--Revise (GVR) Loop}
To enforce these constraints dynamically, we deploy a bounded iterative refinement mechanism (see Algorithm~\ref{alg:inference}). Initially, the fine-tuned model $M_\Theta$ generates a preliminary S-CoT response $P^{(0)}$. This tentative proposal is immediately parsed by the safety module to compute the violation set $V^{(0)}$ via Eq.~\eqref{eq:violation_set}. Crucially, if violations are detected, the system does not simply reject the output; instead, it injects a Negative Feedback Signal. A corrective prompt, enriched with specific evidence triples (e.g., ``Error: LI4 is ProhibitedFor Pregnancy''), is fed back into the model. This explicit error signal guides the model to revise the prescription while retaining the valid diagnostic logic.

To guarantee safety convergence, this refinement loop is bounded by a maximum iteration threshold $T_{max}$. In scenarios where the model fails to resolve conflicts within these steps (i.e., $V^{(T_{max})} \neq \emptyset$), a Conservative Fail-Safe is triggered. This Hard Fallback mechanism suppresses the neural output entirely, reverting to a default warning. This design ensures that CORE-Acu operates strictly within an explicit Safety Envelope, preventing the release of hazardous hallucinations.

\begin{algorithm}[t]
  \caption{Closed-Loop Inference with Symbolic Veto}
  \label{alg:inference}
  \DontPrintSemicolon
  \SetAlgoLined

  \KwIn{Query $Q$, Model $M_{\theta}$, KG $\mathcal{G}$, Unsafe Relations $\mathcal{R}_{\text{unsafe}}$, Budget $T_{\max}$}
  \KwOut{Safe Recommendation $P$ or Fallback}

  $\mathcal{C} \leftarrow \textsc{ExtractConditions}(Q)$ \tcp*[r]{Patient Context}
  $Q^{(0)} \leftarrow Q$\;

  \For{$t \leftarrow 0$ \KwTo $T_{\max}-1$}{
    $P^{(t)} \leftarrow M_{\theta}(Q^{(t)})$ \tcp*[r]{Neural Proposal}
    $\mathcal{A}^{(t)} \leftarrow \textsc{ExtractAcupoints}(P^{(t)})$\;
    $V^{(t)} \leftarrow \textsc{CheckViolations}(\mathcal{C},\mathcal{A}^{(t)},\mathcal{G},\mathcal{R}_{\text{unsafe}})$ \tcp*[r]{Symbolic Veto}
    
    \If{$V^{(t)}=\emptyset$}{
      \Return{$P^{(t)}$} \tcp*[r]{Accepted}
    }
    \tcp{Inject Negative Feedback}
    $Q^{(t+1)} \leftarrow \textsc{ConstructFeedback}(Q, P^{(t)}, V^{(t)})$\;
  }
  \Return{\textsc{SafeFallback}(\text{``Human Confirmation Required''})}\;
\end{algorithm}

\section{Experiments}
In this section, we present a comprehensive evaluation of the CORE-Acu framework, focusing on reasoning trace quality, safety compliance, and entity fidelity. We first outline the datasets and baseline models, followed by the evaluation protocols. For reproducibility, the detailed implementation specifications, including hardware configurations and hyperparameter settings, are summarized in Table~\ref{tab:impl_details}.
\subsection{Datasets}
\label{sec:datasets}
We evaluate CORE-Acu using the Acu-Reasoning dataset (constructed in Section~\ref{sec:scot_dataset}) and a strictly held-out test set. The data partitions are defined as follows:

\begin{itemize}
    \item \textbf{Training and Validation Sets:} 
    The Acu-Reasoning corpus comprises a total of 42,512 samples. We partition this corpus into a training set of 38,260 samples (90\%) and a validation set of 4,252 samples (10\%) for hyperparameter tuning and early stopping. This split ensures that the model learns the structured S-CoT schema and domain logic prior to final evaluation.

    \item \textbf{Held-out Test Set (Quantitative, $N=1,000$):} 
    To assess generalization, we utilize 1,000 de-identified clinical cases that are strictly excluded from the training and validation phases. These cases cover diverse departments and disease types, serving as the basis for large-sample quantitative evaluation metrics (Surface Overlap, Safety, and LLM-as-a-Judge scoring).

    \item \textbf{Complex Case Subset (Qualitative, $n=4$):} 
    From the held-out test set, we select four clinically representative complex cases. These cases are characterized by conflicting symptoms or multi-pattern complications, serving as a focused subset for fine-grained human expert review to inspect reasoning coherence and interpretability.
\end{itemize}

\begin{table}[htbp]
\centering
\caption{Implementation Details of Experimental Setup.}
\label{tab:impl_details}
\small
\renewcommand{\arraystretch}{0.95}

\begin{tabularx}{\linewidth}{l X}
    \toprule
    \textbf{Parameter} & \textbf{Value / Specification} \\
    \midrule

    \multicolumn{2}{l}{\textit{\textbf{Hardware Infrastructure}}} \\
    \quad CPU & 16 vCPU on AMD EPYC 9654 \\
    \quad GPU & NVIDIA GeForce RTX 4090 (24GB) \\
    \midrule

    \multicolumn{2}{l}{\textit{\textbf{LoRA Fine-tuning Setup}}} \\
    \quad Base Model & Qwen3-8B \\
    \quad Fine-tuning Type & LoRA \\
    \quad Rank $r$ & 16 \\
    \quad LoRA Scaling & 32 \\
    \quad LoRA Dropout & 0.05 \\
    \quad Max Seq. Length & 2048 \\
    \quad Epochs & 3 \\
    \quad Learning Rate & $1 \times 10^{-4}$ \\
    \quad Warmup Ratio & 0.05 \\
    \quad Batch Size (Per-device) & 2 \\
    \quad Grad. Accum. Steps & 16 (Effective batch = 32) \\
    \quad Weight Decay & 0.01 \\
    \quad Gradient Clipping & Max norm 0.5 \\
    \quad Validation Split & 10\% held out \\
    \quad Keyword Weight & 1.5 \\
    \midrule

    \multicolumn{2}{l}{\textit{\textbf{KG-Constrained Inference}}} \\
    \quad Retry Budget $T_{\max}$ & 5 \\
    \quad Unsafe Relations & \texttt{Incompatible}, \texttt{ProhibitedFor} \\
    \quad Fallback Policy & Human confirmation request \\
    \bottomrule
\end{tabularx}
\end{table}

\subsection{Baselines}
We benchmark CORE-Acu against five baselines spanning frontier general-purpose LLMs and medical-domain models, so as to assess both overall generation quality and the ability to comply with acupuncture-specific clinical constraints.
\begin{itemize}
    \item \textbf{GPT-4o:} A frontier proprietary general-purpose LLM released by OpenAI, widely used as a strong reference point for general reasoning and instruction following. 
    We include it to represent a high-capability closed-source model and to examine how far pure probabilistic generation can go without domain-specific safety governance.

    \item \textbf{Qwen-Max:} A flagship commercial Chinese general-purpose LLM in the Qwen family, available via API and designed to support long-context usage in practical applications. 
    We select it as a strong Chinese-language baseline to control for language coverage and to benchmark against a widely deployed industrial model.

    \item \textbf{GLM-4.6:} A recent general-purpose model from Zhipu AI that is positioned for complex agentic and coding-oriented workloads and supports a long context window. 
    We use it as an additional strong non-medical baseline to test whether improvements in general capability and long-context handling translate into safer acupuncture prescriptions without explicit clinical constraints.

    \item \textbf{HuatuoGPT-o1-7B:} An open medical LLM oriented toward medical reasoning, which explicitly produces a deliberative reasoning process before the final answer. 
    We include it to evaluate whether domain-specialized medical models better respect acupuncture-related clinical constraints compared with general-purpose LLMs.

    \item \textbf{Qwen3-8B (base):} The pretrained backbone model used in our framework prior to any domain adaptation or instruction tuning. 
    This baseline isolates the contribution of our fine-tuning and entity-aware optimization by quantifying improvements over the same underlying base model.
\end{itemize}

\subsection{Evaluation Metrics and Protocols}
\label{sec:metrics}

To comprehensively assess model performance, we design a hierarchical evaluation strategy organized into three distinct levels of scrutiny: (1) Automated Surface Metrics to measure lexical fidelity and entity coverage; (2) KG-Grounded Verification to enforce hard safety boundaries; and (3) a Hybrid Multidimensional Assessment to evaluate clinical reasoning depth.

\begin{enumerate}
    \item \textbf{Surface Overlap and Entity Fidelity:}
    We report BLEU-4 and ROUGE-L to quantify surface-form lexical proximity to reference prescriptions. All outputs are pre-processed by removing formatting artifacts and tokenized using Jieba for Chinese text.
    
    \quad BLEU-4 is computed at the sentence level with smoothing and macro-averaged over $N$ cases:
    \begin{equation}
    \label{eq:bleu4_impl}
    \mathrm{BLEU}\text{-}4=\frac{1}{N}\sum_{i=1}^{N}\mathrm{BP}^{(i)}\cdot
    \exp\!\left(\frac{1}{4}\sum_{n=1}^{4}\log \tilde{p}^{(i)}_{n}\right)
    \end{equation}
    where $\tilde{p}^{(i)}_{n}$ denotes the smoothed modified $n$-gram precision and $\mathrm{BP}^{(i)}$ is the brevity penalty.
    
    \quad ROUGE-L (F1) is computed based on the longest common subsequence (LCS):
    \begin{equation}
    \label{eq:rougel_impl}
    \mathrm{ROUGE}\text{-}L=\frac{1}{N}\sum_{i=1}^{N}
    \frac{2P^{(i)}_{\mathrm{LCS}}R^{(i)}_{\mathrm{LCS}}}{P^{(i)}_{\mathrm{LCS}}+R^{(i)}_{\mathrm{LCS}}}
    \end{equation}
    
    \quad To assess domain-specific precision beyond general text overlap, we define \textit{Entity-F1}. We extract key medical entities (POS tags: \texttt{n, nr, ns, nt, nz, vn}; length $>1$) and compute the macro-averaged F1 score:
    \begin{equation}
    \label{eq:entityf1_impl}
    \mathrm{Entity}\text{-}\mathrm{F1}=\frac{1}{N}\sum_{i=1}^{N}
    \frac{2P^{(i)}_{\mathrm{ent}}R^{(i)}_{\mathrm{ent}}}{P^{(i)}_{\mathrm{ent}}+R^{(i)}_{\mathrm{ent}}}
    \end{equation}
    where $P^{(i)}_{\mathrm{ent}}$ and $R^{(i)}_{\mathrm{ent}}$ denote the precision and recall of the predicted entity set $E^{(i)}_{\mathrm{pred}}$ against the reference $E^{(i)}_{\mathrm{ref}}$.

    \item \textbf{KG-Grounded Safety Verification:}
    Since overlap metrics cannot detect dangerous medical errors, we employ our TCM-KG to compute two rigorous safety indicators:
    
    \textit{Entity Hallucination Rate (EntHall):} This measures the proportion of generated acupoints that do not exist in the canonical ontology. For case $i$, with predicted acupoint set $\mathcal{A}^{(i)}_{\text{pred}}$ and KG vocabulary $\mathcal{A}_{\text{KG}}$:
    \begin{equation}
    \label{eq:enthall_case}
    \mathrm{EntHall}=\frac{1}{N}\sum_{i=1}^{N}
    \frac{\left|\mathcal{A}^{(i)}_{\text{pred}}\setminus \mathcal{A}_{\text{KG}}\right|}
    {\left|\mathcal{A}^{(i)}_{\text{pred}}\right|+\epsilon}
    \end{equation}
    
    \textit{Safety Violation Rate (SVR):} This metric flags strictly prohibited recommendations. A case is deemed unsafe if the verifier detects any evidence triple in the generated output. Let $V(\mathcal{C}^{(i)}, \mathcal{A}^{(i)}_{\text{pred}})$ be the violation set returned by the safety module:
    \begin{equation}
    \label{eq:svr}
    \mathrm{SVR}=\frac{1}{N}\sum_{i=1}^{N}\mathbb{I}\!\left[V(\mathcal{C}^{(i)},\mathcal{A}^{(i)}_{\text{pred}})\neq \emptyset\right]\times 100\%
    \end{equation}

    \item \textbf{Multidimensional Reasoning Assessment (Hybrid Protocol):}
    To capture the nuances of clinical logic missed by automated metrics while balancing evaluation interpretability (depth) with coverage (breadth), we evaluate reasoning quality using a five-dimension clinical rubric (0--10 points each, 0--50 total):
    \begin{itemize}
        \item[] \textit{Reasoning Validity:} Logical coherence of the causal chain.
        \item[] \textit{Diagnostic Accuracy:} Correctness of syndrome differentiation.
        \item[] \textit{Pathological Logic:} Alignment with TCM etiological theory.
        \item[] \textit{Prescription Rationality:} Justification of acupoint selection.
        \item[] \textit{Explicability:} Structural clarity and professional standardization.
    \end{itemize}
    
    \noindent\textbf{Protocol:} We implement this rubric via a Hybrid Approach:
    \begin{itemize}
        \item \textit{Depth (Human Expert):} Three senior TCM practitioners ($>10$ years experience) perform an independent, back-to-back assessment of the Complex Case Subset ($n=4$). To guarantee objectivity and mitigate consensus bias, experts evaluate the interpretability of the generated content in strictly isolated environments, serving as a rigorous benchmark for clinical validity.
        \item \textit{Breadth (LLM-as-a-Judge):} GPT-5.2 evaluates the full Held-out Test Set ($N=1,000$) to statistically validate the scalability of reasoning improvements.
    \end{itemize}
\end{enumerate}

\begin{table}[htbp]
    \centering
    \caption{Detailed Configuration of the LLM-as-a-Judge Evaluation.}
    \label{tab:judge_details}

    \renewcommand{\arraystretch}{1.2}
    \small 
    
    \begin{tabularx}{\columnwidth}{l X}
        \toprule
        \textbf{Configuration Item} & \textbf{Value / Specification} \\
        \midrule
        Judge Model & GPT-5.2 (via OpenAI API) \\
        Call Date & December 2025 \\
        Temperature & 0.1 \\
        Top\_p & 1.0 \\
        Max Tokens & 2048 \\
        
        Scoring Scale & 0--50 Total Score \newline (5 dimensions, 0--10 each) \\
        
        Evaluated Cases & 1{,}000 clinical cases \\
        \bottomrule
    \end{tabularx}
\end{table}

\section{Main Results}
In this section, we systematically evaluate the performance of CORE-Acu across diverse clinical dimensions. We begin by reporting the automated surface and entity metrics to assess lexical fidelity (Section~\ref{sec:exp_surface}). Subsequently, we present a multidimensional reasoning assessment using the LLM-as-a-judge protocol, followed by a sensitivity analysis of the entity re-weighting factor. Finally, we examine the framework's safety compliance through KG-grounded verification and corroborate the findings with qualitative expert review.

\subsection{Textual Generation Quality: Lexical Fidelity and Entity Precision}
\label{sec:exp_surface}
To assess the generation quality on a textual level, we conducted an automated evaluation on the held-out test set comprising 1{,}000 cases. Table~\ref{tab:auto_eval} summarizes the performance of CORE-Acu against the baseline models across three representative metrics: BLEU-4, ROUGE-L, and the domain-specific Entity-F1 Score. Detailed breakdowns for all n-gram and entity metrics (including BLEU-1/2/3, ROUGE-1/2, and Precision/Recall) are provided in~\ref{app:results}.

\begin{table}[htbp]
\centering
\caption{\textbf{Automated Evaluation Results on the Held-out Test Set.} CORE-Acu achieves the best performance among the evaluated baselines across all metrics. Note: \textit{Entity-F1} measures the strict overlap of key medical entities (Acupoints, Syndromes) against the ground truth.}
\label{tab:auto_eval}
\renewcommand{\arraystretch}{1.15}
\setlength{\tabcolsep}{4pt}

\resizebox{\columnwidth}{!}{%
\begin{tabular}{lccc}
\toprule[1.5pt]
\textbf{Model} & \textbf{BLEU-4} & \textbf{ROUGE-L} & \textbf{Entity-F1} \\
\midrule
\textbf{CORE-Acu (Ours)} & \textbf{0.1898} & \textbf{0.3334} & \textbf{0.4612} \\
GLM-4.6                  & 0.0697          & 0.1970          & 0.3511          \\
HuatuoGPT-o1-7B    & 0.0059          & 0.1209          & 0.2161          \\
Qwen3-8B (Base)          & 0.0814          & 0.1883          & 0.3062          \\
Qwen-Max (General)       & 0.0865          & 0.1988          & 0.3087          \\
GPT-4o             & 0.0722          & 0.1985          & 0.2906          \\
\bottomrule[1.5pt]
\end{tabular}%
}
\end{table}

As indicated in Table~\ref{tab:auto_eval}, CORE-Acu demonstrates a clear advantage in domain-specific fidelity. Among the general-purpose baselines (GPT-4o, Qwen-Max, and GLM-4.6), the generated text is often linguistically fluent, yet their lower Entity-F1 scores (0.2906, 0.3087, and 0.3511, respectively) suggest that they still tend to hallucinate, substitute, or omit clinically salient entities such as acupoints and syndromic terms. In contrast, CORE-Acu attains an Entity-F1 of 0.4612, supporting the effectiveness of the LMERL (Section~\ref{sec:lmerl}) in improving attention to hard medical facts and reducing entity-level inconsistency.

The medical-domain baseline HuatuoGPT-o1-7B underperforms across all three metrics (BLEU-4: 0.0059; ROUGE-L: 0.1209; Entity-F1: 0.2161). One likely factor is that, while HuatuoGPT is optimized for general medical dialogue, it does not explicitly enforce the task-specific structured output and acupuncture entity constraints required by our evaluation protocol; consequently, its responses may deviate in format and omit key standardized entities, leading to poor surface overlap and strict entity matching.

Notably, despite not optimizing directly for surface-level overlap, CORE-Acu also leads in BLEU-4 and ROUGE-L. This may be driven by the S-CoT structure (Section~\ref{sec:scot_dataset}), which aligns the model's output schema more strictly with the standardized references used in the test set. Conversely, baseline models may produce verbose, paraphrased, or weakly structured responses that drift from the gold standard, resulting in lower n-gram overlap scores.

\subsection{Reasoning Competence: Logical Coherence and Multidimensional Validity}
\begin{figure}[htbp]
  \centering
  \includegraphics[width=1.0\linewidth]{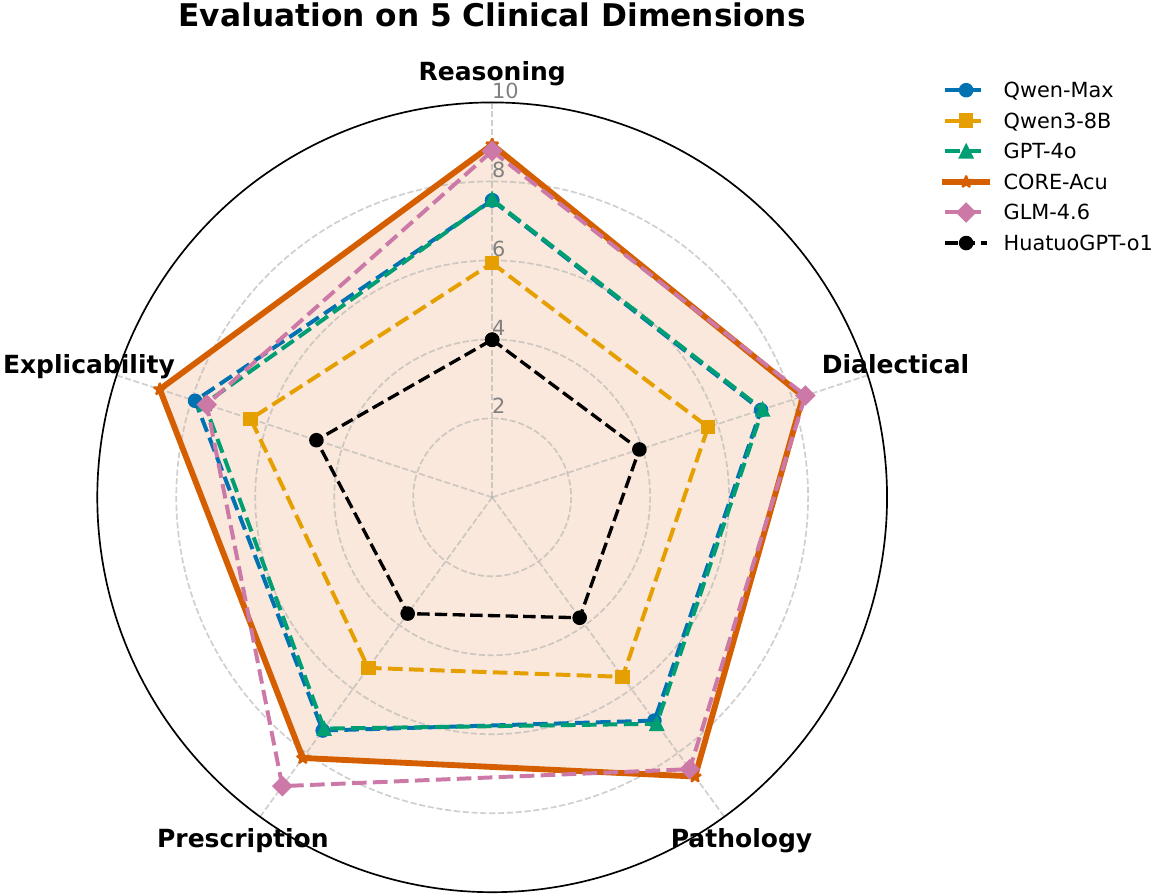}
  \caption{\textbf{Comparative performance across five clinical dimensions.}}
  \label{fig:radar_chart}
\end{figure}

\begin{figure}[htbp]
  \centering
  \includegraphics[width=1.0\linewidth]{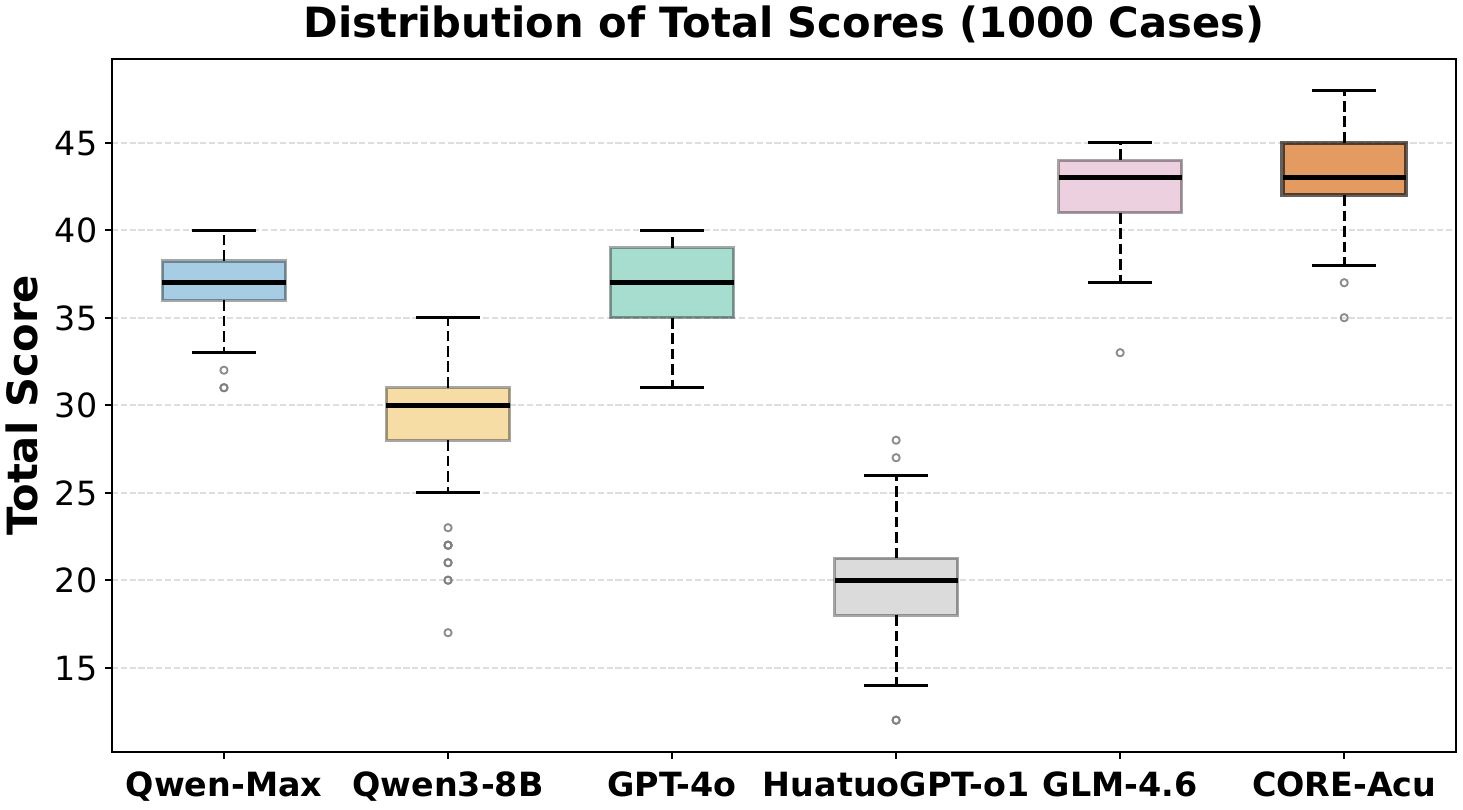}
  \caption{\textbf{Distribution of total scores across 1{,}000 test cases.}
  The comparison includes CORE-Acu, Qwen-Max, Qwen3-8B, GPT-4o, GLM-4.6, and HuatuoGPT-o1-7B.}
  \label{fig:boxplot_total_score}
\end{figure}

Given that n-gram overlap metrics are limited in capturing logical coherence and causal validity, we used GPT-5.2 as an automated judge to evaluate the 1{,}000 test cases along the five dimensions defined in Section~\ref{sec:metrics}, with detailed configuration provided in Table~\ref{tab:judge_details}. Figure~\ref{fig:radar_chart} summarizes the dimension-wise comparison with a radar chart, and Figure~\ref{fig:boxplot_total_score} reports the distribution of total scores.

% -------------------------------------------------------
% Boxplot
% -------------------------------------------------------

The radar chart highlights a critical ``reasoning gap'' in baseline models. Although strong general-purpose LLMs can be competitive on Prescription Rationality, they underperform more noticeably on reasoning-centric dimensions such as Dialectical Accuracy and Pathological Logic. This pattern suggests that a general-purpose model may arrive at a plausible prescription without consistently deriving it from correct syndrome differentiation and a coherent pathological mechanism.
Meanwhile, the medical/TCM-oriented baseline (HuatuoGPT-o1-7B) does not necessarily translate domain fluency into acupuncture-specific causal alignment, reinforcing the need for task-structured supervision.
By contrast, CORE-Acu maintains high scores across all axes and is especially strong in Reasoning Validity and Pathological Logic, indicating that S-CoT fine-tuning effectively instills the intended syndrome-to-Treatment causal logic and improves the soundness of the decision process. The full judge prompt and scoring rubric are provided in~\ref{app:judge}.

\subsection{Optimization Dynamics: Balancing Linguistic Fluency with Entity Importance}
\label{sec:sensitivity_alpha}

We conduct a sensitivity analysis on the internal validation set to determine the optimal entity weight factor $\alpha$ for the LMERL objective. The baseline setting $\alpha=1.0$ corresponds to standard supervised fine-tuning with uniform cross-entropy. While absolute validation scores naturally exceed those on the held-out test set due to distribution shifts, the relative performance trends remain consistent, justifying the use of this set for hyperparameter selection.

As presented in Table~\ref{tab:ablation_alpha}, setting $\alpha=1.5$ yields the optimal trade-off. Compared to the baseline, this configuration achieves a 3.2\% relative gain in Entity-F1 ($0.5350 \to 0.5521$) while simultaneously improving surface-form metrics (BLEU-4 and ROUGE-L). This indicates that moderate gradient upweighting enhances domain entity precision without compromising linguistic fluency. Conversely, increasing $\alpha$ beyond 1.5 leads to performance degradation across all metrics. Specifically, at $\alpha=3.0$, both Entity-F1 and Entity-Recall drop significantly ($0.5034$ and $0.4806$, respectively), suggesting that excessive bias toward keywords disrupts generalization and semantic coherence. Consequently, we fix $\alpha=1.5$ for all subsequent experiments.

\begin{table}[!htbp]
  \centering
  \caption{Sensitivity analysis of the keyword weight factor $\alpha$ on the internal validation set.}
  \label{tab:ablation_alpha}
  \setlength{\tabcolsep}{3pt}
  \resizebox{\linewidth}{!}{
    \begin{tabular}{lccccc}
      \toprule
      \textbf{$\alpha$} & \textbf{BLEU-4} & \textbf{ROUGE-L} & \textbf{Entity-P} & \textbf{Entity-R} & \textbf{Entity-F1} \\
      \midrule
      1.0 (Baseline) & 0.3728 & 0.4452 & 0.5492 & 0.5245 & 0.5350 \\
      1.02 & 0.3715 & 0.4453 & 0.5495 & 0.5284 & 0.5368 \\
      \textbf{1.5 (Ours)} & \textbf{0.3879} & \textbf{0.4573} & \textbf{0.5667} & \textbf{0.5425} & \textbf{0.5521} \\
      2.0 & 0.3485 & 0.4225 & 0.5430 & 0.4937 & 0.5139 \\
      3.0 & 0.3455 & 0.4206 & 0.5332 & 0.4806 & 0.5034 \\
      \bottomrule
    \end{tabular}
  }
\end{table}

\subsection{Clinical Reasoning Evaluation: Qualitative Depth and Quantitative Validation}
\label{sec:clinical_eval}
To rigorously validate whether the model's generated reasoning traces reflect genuine clinical logic rather than stochastic mimicry, we present a dual-perspective analysis. This section reports findings from both fine-grained expert review (to assess reasoning depth and logical soundness) and large-scale automated validation (to verify the generalizability of these capabilities).

\subsubsection{Qualitative Expert Review (Depth)}
The consensus results from expert review are summarized in Table~\ref{tab:expert_eval}. Given the intentionally small sample size ($n=4$), we do not treat this component as a hypothesis test; instead, we use it to surface concrete failure modes and to provide clinically grounded interpretations of model behavior.

\begin{table*}[htbp]
\centering
\caption{\textbf{Qualitative clinical expert evaluation (consensus scores, 0--10; $n=4$ complex cases).}
Three senior TCM practitioners assessed each model along five criteria. GPT-4o attains the highest \textit{Prescription} score, whereas CORE-Acu achieves substantially stronger \textit{Pathology Logic} and \textit{Explicability}, indicating better alignment between clinical reasoning and the final prescription.}
\label{tab:expert_eval}
\renewcommand{\arraystretch}{1.3}
\setlength{\tabcolsep}{14pt}

\begin{tabular}{lccccc}
\toprule[1.5pt]
\textbf{Model} & \textbf{Reasoning} & \textbf{Diagnosis} & \textbf{Pathology Logic} & \textbf{Prescription} & \textbf{Explicability} \\
\midrule
\textbf{CORE-Acu (Ours)} & \textbf{9.53} & \textbf{9.33} & \textbf{8.00} & 8.00 & \textbf{10.00} \\
GPT-4o & 6.00 & 2.00 & 2.00 & \textbf{10.00} & 7.00 \\
GLM-4.6 & 6.33 & 6.00 & 6.00 & 6.33 & 7.87 \\
HuatuoGPT & 4.00 & 3.00 & 3.33 & 7.00 & 5.67 \\
Qwen-Max & 6.00 & 6.00 & 2.00 & 8.00 & 2.00 \\
Qwen3-8B & 5.00 & 2.00 & 2.00 & 8.00 & 7.00 \\
\bottomrule[1.5pt]
\end{tabular}
\end{table*}

\begin{table*}[t]
\centering
\caption{\textbf{Quantitative scale validation on 1{,}000 clinical cases.}
Scores are reported as mean $\pm$ SD over 1{,}000 cases using the LLM-as-a-Judge protocol (GPT-5.2). CORE-Acu attains the best overall performance and maintains consistent gains on reasoning-centric dimensions.}
\label{tab:llm_scale_eval}
\renewcommand{\arraystretch}{1.2}
\setlength{\tabcolsep}{8pt}

\begin{tabular*}{\textwidth}{@{\extracolsep{\fill}}lcccccc}
\toprule
\textbf{Model} & \textbf{Total Score} & \textbf{Reasoning} & \textbf{Dialectical} & \textbf{Pathology Logic} & \textbf{Prescription} & \textbf{Explicability} \\
\midrule
GPT-4o            & 36.73 $\pm$ 2.49 & 7.54 $\pm$ 0.56 & 7.20 $\pm$ 0.65 & 7.09 $\pm$ 0.58 & 7.24 $\pm$ 0.63 & 7.66 $\pm$ 0.47 \\
Qwen-Max          & 36.88 $\pm$ 2.10 & 7.52 $\pm$ 0.54 & 7.16 $\pm$ 0.58 & 6.99 $\pm$ 0.59 & 7.30 $\pm$ 0.56 & 7.91 $\pm$ 0.32 \\
Qwen3-8B          & 29.06 $\pm$ 3.56 & 5.93 $\pm$ 0.79 & 5.75 $\pm$ 0.77 & 5.62 $\pm$ 0.72 & 5.34 $\pm$ 1.14 & 6.43 $\pm$ 0.59 \\
GLM-4.6           & 42.26 $\pm$ 2.12 & 8.77 $\pm$ 0.44 & 8.34 $\pm$ 0.60 & 8.51 $\pm$ 0.62 & \textbf{ 9.04 }$\pm$ 0.34 & 7.60 $\pm$ 0.58 \\
HuatuoGPT            & 20.00 $\pm$ 3.15 & 3.99 $\pm$ 0.76 & 3.92 $\pm$ 0.67 & 3.77 $\pm$ 0.69 & 3.64 $\pm$ 0.83 & 4.68 $\pm$ 0.73 \\
\textbf{CORE-Acu} & \textbf{42.96 $\pm$ 2.47} & \textbf{8.93 $\pm$ 0.59} & \textbf{8.28 $\pm$ 0.65} & \textbf{8.74 $\pm$ 0.52} & 8.16 $\pm$ 0.72 & \textbf{8.85 $\pm$ 0.55} \\
\bottomrule
\end{tabular*}
\end{table*}

Table~\ref{tab:expert_eval} reveals a pronounced mismatch between prescription-level quality and reasoning-level validity in several baselines.
Memorization trap. GPT-4o achieves a perfect \textit{Prescription} score (10.00), yet its \textit{Diagnosis} and \textit{Pathology Logic} scores drop to 2.00. This pattern suggests an ``answer-first'' behavior: the model can generate a plausible (or even correct) prescription while failing to consistently justify it through syndrome differentiation and coherent pathological reasoning, which limits interpretability in clinical decision support.
CORE-Acu's reasoning alignment. In contrast, CORE-Acu exhibits a substantially more balanced profile. It improves \textit{Pathology Logic} to 8.00 (vs.\ 2.00 for GPT-4o) and achieves the highest \textit{Explicability} (10.00). These results are consistent with the intended effect of S-CoT training: strengthening the linkage between syndrome-level reasoning and the resulting acupoint prescription, thereby reducing ungrounded, black-box outputs.

\subsubsection{Quantitative Scale Validation Results}
Quantitative scoring via the LLM-as-a-Judge protocol on the 1,000 held-out cases (Table~\ref{tab:llm_scale_eval}) confirms that CORE-Acu achieves the highest comprehensive performance among all baselines. Crucially, the model maintains substantial margins in \textit{Reasoning Validity} ($8.93 \pm 0.59$) and \textit{Pathological Logic} ($8.74 \pm 0.52$). 

This consistency across a large-scale evaluation corroborates the findings of the expert review: CORE-Acu's performance gains stem from the generation of coherent diagnostic chains and syndrome-grounded theoretical rationales, rather than merely superficial textual optimization. Although automated scoring cannot fully replace prospective clinical trials, these results provide strong convergent evidence that the framework's reasoning advantages possess robust generalization capabilities, extending well beyond specific anecdotal cases.

\subsection{Safety Assurance: Enforcing Hard Constraints via Symbolic Verification}
\begin{figure}[htbp]
    \centering
    \includegraphics[width=0.95\linewidth]{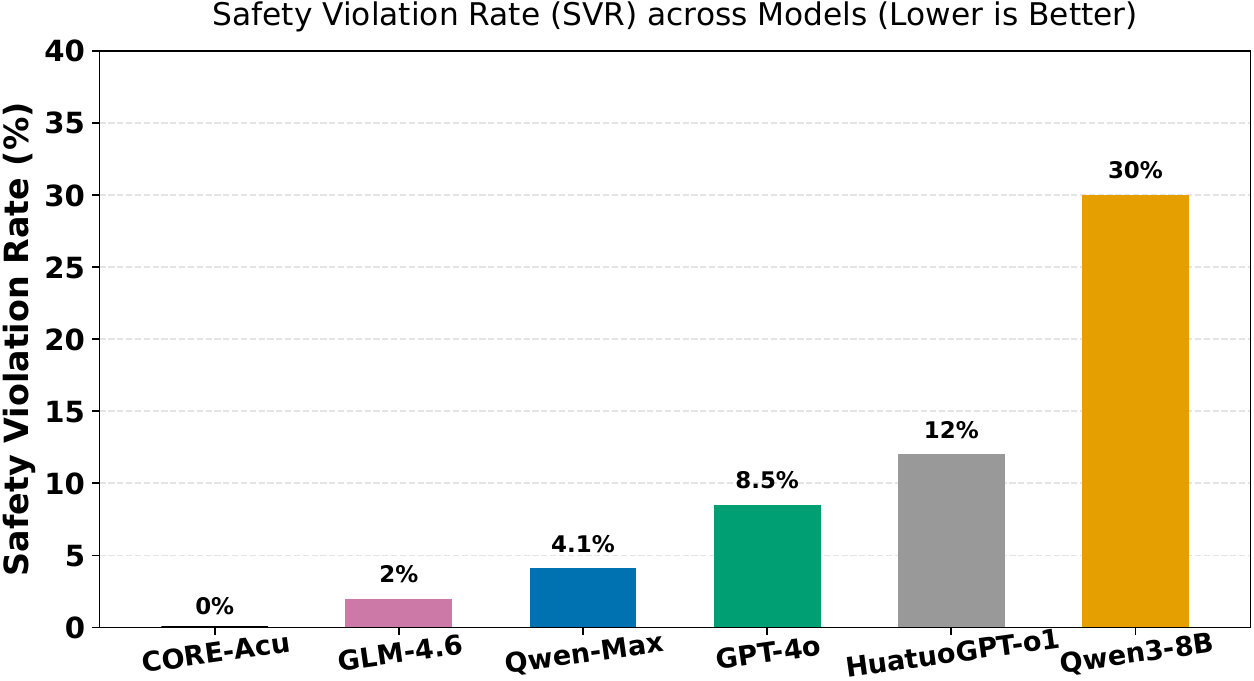}
    \caption{\textbf{Safety Violation Rate across models (lower is better).}}
    \label{fig:safety_svr}
\end{figure}

To quantify safety, we report the SVR as defined in Eq.~\eqref{eq:svr}. Concretely, a case is counted as unsafe if the KG-based verifier returns at least one evidence triple indicating a contraindication or incompatibility under the KG-encoded rule set. 
Violations include (i) acupoint combinations explicitly prohibited by acupuncture safety rules (e.g., pregnancy-related restrictions such as \textit{Hegu} + \textit{Sanyinjiao}) and (ii) condition--acupoint conflicts triggered by patient-specific conditions extracted from the query. 
Figure~\ref{fig:safety_svr} reports SVR across models.

Figure~\ref{fig:safety_svr} indicates that unconstrained foundation models may generate clinically unsafe prescriptions under KG-grounded rule-based evaluation. Despite strong linguistic fluency, GPT-4o produces contraindicated acupoint combinations in 8.5\% of test cases, and Qwen-Max shows 4.1\% SVR. The additional general-purpose baseline GLM-4.6 still exhibits an SVR of 2.0\%, suggesting that scaling or alternative general pre-training alone does not reliably enforce hard acupuncture safety constraints.
Notably, the medical/TCM-oriented baseline HuatuoGPT-o1-7B incurs 12.0\% SVR, reinforcing that domain-fluent medical dialogue capability does not automatically translate into acupuncture-specific contraindication compliance without explicit constraint mechanisms.

In contrast, CORE-Acu attains an \emph{observed} SVR of 0/1{,}000 on the held-out test set under the safety relations explicitly encoded in our KG. The improvement is attributable to the KG-Constrained Inference mechanism (Section~\ref{sec:kg_infer}), which performs post-generation verification and triggers violation-aware rewriting in a bounded loop. If conflicts persist after the retry budget is exhausted, CORE-Acu returns a conservative fallback requesting human confirmation, thereby reducing the likelihood of emitting prescriptions that violate KG-defined safety rules.

To isolate the contribution of the symbolic safety module, we conduct an ablation study by disabling the post-hoc KG verification and rewrite loop. Evaluation is performed on 1{,}000 held-out test cases. Although these cases are strictly excluded from training to reduce the risk of data leakage, all entities appearing in the test set are covered by the KG ontology, so that rule-based safety constraints remain verifiable at inference time. This design tests generalization to unseen patient descriptions while ensuring that safety violations, if present, are detectable under the KG-defined rule set.

\begin{table}[htbp]
  \centering
  \small
  \setlength{\tabcolsep}{2.5pt}
  \renewcommand{\arraystretch}{1.08}
  \caption{Ablation of KG veto (1{,}000 test cases).}
  \label{tab:ablation_veto}

  \begin{tabularx}{\columnwidth}{@{}
    >{\raggedright\arraybackslash}p{0.20\columnwidth}
    >{\centering\arraybackslash\hspace{0.35em}}p{0.18\columnwidth} % <-- add right padding + slightly wider
    >{\centering\arraybackslash}p{0.10\columnwidth}
    >{\centering\arraybackslash}p{0.31\columnwidth} % <-- slightly narrower
    >{\centering\arraybackslash}p{0.12\columnwidth}
  @{}}
    \toprule
    \textbf{Method} &
    \textbf{Mech.} &
    \makecell{\textbf{SVR}\\\textbf{(\%)}} &
    \makecell{\textbf{Repair}\\\textbf{(fix/det)}} &
    \makecell{\textbf{FB}\\\textbf{(\#)}} \\
    \midrule
    \makecell[tl]{Qwen3-8B\\(base)} & -- & 30.0 & -- & -- \\
    \makecell[tl]{CORE-Acu\\(SFT)}  & -- & 4.0  & -- & -- \\
    \makecell[tl]{\textbf{CORE-Acu}\\\textbf{(full)}} &
    \textbf{veto\,{+}\,rw} & \textbf{0 (0/1{,}000)} & \textbf{100\% (40/40)} & \textbf{0} \\
    \bottomrule
  \end{tabularx}
\end{table}

As shown in Table~\ref{tab:ablation_veto}, the \textit{SFT-only} variant (trained with the optimized keyword weight $\alpha=1.5$) reduces SVR from 30.0\% (base model) to 4.0\%. This indicates that fine-tuning internalizes a substantial portion of explicit contraindication patterns. Nevertheless, the remaining 4.0\% SVR is still material for high-stakes clinical settings, motivating the need for an explicit constraint enforcement layer.

Re-enabling KG-Constrained Inference (CORE-Acu full) reduces the \emph{observed} Safety Violation Rate to 0/1{,}000 under our KG-encoded rule set by verifying first-pass outputs and triggering bounded revisions when violations are found. Across the 1{,}000 cases, the verifier detected 40 violations in the first-pass generations.
A violation was counted as \emph{repaired} if the rewritten prescription removed all KG-matched unsafe relations in $\mathcal{R}_{\text{unsafe}}$ within $T_{\max}=5$ attempts. 
Under this criterion, all detected violations were repaired within the retry budget (40/40), yielding an observed SVR of 0/1{,}000 on this test set. Fallback was available but not triggered on this benchmark.
These results suggest that the model retains sufficient flexibility to produce alternative prescriptions when guided by symbolic negative feedback.

\section{Discussion}
\label{sec:discussion}

Based on systematic evaluations comprising automated metrics, knowledge-graph-based verification, and multi-dimensional expert adjudication, we conclude that the proposed CORE-Acu framework addresses critical challenges in high-stakes clinical decision support through three defining properties: (1) Traceable auditability of clinical logic — By introducing an explicit S-CoT schema (symptom $\rightarrow$ diagnosis $\rightarrow$ pathology $\rightarrow$ principle $\rightarrow$ prescription), CORE-Acu transforms implicit TCM inference into observable intermediate states. This explicit causal modeling significantly reduces the reasoning--outcome mismatch frequently observed in general-purpose LLMs, ensuring that clinical decisions are not merely probabilistic outputs but result from verifiable diagnostic logic that aligns with expert cognitive patterns. (2) Symbolic Veto Mechanism for Hard Constraint Enforcement — Unlike standard probabilistic generation, our framework effectively eliminates hard constraint violations through a KG-grounded verification mechanism. The implementation of a \textit{Generate--Verify--Revise} inference loop provides deterministic intercept capabilities against contraindications (e.g., pregnancy prohibitions), utilizing bounded rewriting and conservative fallbacks to strictly maintain a zero-violation safety envelope when conflicts persist. (3) Mitigating Frequency-Importance Mismatch via LMERL — The framework addresses the intrinsic frequency--importance mismatch in medical corpus training through the LMERL. This optimization strategy adaptively up-weights gradients for safety-critical domain terms, successfully mitigating terminology drift and ensuring high-precision handling of acupoint names and specific syndromes without relying solely on prompt engineering.

\section{Conclusion}
\label{sec:conclusion}

In this paper, we propose CORE-Acu, a neuro-symbolic framework for acupuncture clinical decision support designed to address two critical challenges in LLM deployment: opaque reasoning and safety-critical hallucinations. We first introduce a schema-constrained S-CoT supervision paradigm to transform implicit TCM logic into an explicit causal chain, thereby resolving the black-box issue of clinical reasoning. To mitigate terminology drift, we propose the LMERL objective to adaptively reweight gradients during fine-tuning, ensuring the high-fidelity generation of safety-critical entities. Finally, we implement a KG-grounded symbolic veto mechanism that enforces deterministic contraindication rules through a \textit{Generate--Verify--Revise} loop, providing the rigorous safety guarantees necessary for invasive acupuncture interventions.

Extensive experiments demonstrate the superiority of CORE-Acu across three complementary evaluation mechanisms: automated surface metrics, KG-grounded safety verification, and expert-level multidimensional adjudication. Results show that our framework significantly outperforms general-purpose and medical-domain baselines in both reasoning interpretability and safety compliance, achieving a zero-violation rate on the held-out test set.

Despite these advancements, several challenges remain. Our current knowledge graph coverage is not exhaustive, and the binary veto mechanism may overlook nuanced, context-dependent clinical trade-offs. Future work will focus on extending the KG to encompass a broader range of contraindications, developing graded risk modeling for more flexible decision support, and incorporating multimodal signals—such as tongue and facial images—to enhance evidence grounding.

\section*{Declaration of generative AI and AI-assisted technologies in the writing process}
During the preparation of this work, the author(s) used ChatGPT in order to improve language clarity. After using this tool, the author(s) reviewed and edited the content as needed and take full responsibility for the content of the publication.

\section*{CRediT authorship contribution statement}
\textbf{Liuyi Xu} (Methodology, Software, Formal analysis, Writing -- original draft); 
\textbf{Yun Guo} (Investigation, Data curation, Writing -- review \& editing); 
\textbf{Ming Chen} (Data curation, Validation); 
\textbf{Zihan Dun} (Software, Visualization); 
\textbf{An-Yang Lu} (Resources, Funding acquisition, Validation); 
\textbf{Shuang Li} (Formal analysis, Validation); 
\textbf{Lijun Liu} (Supervision, Funding acquisition, Resources); 
\textbf{Yining Qian} (Conceptualization, Project administration, Supervision, Writing -- review \& editing).

\section*{Ethics approval and consent to participate}
Not applicable. All clinical records were fully de-identified prior to analysis, and no identifiable
private information was accessed. In accordance with local regulations and institutional policies,
ethical review and informed consent were not required for this retrospective, minimal-risk study.

\section*{Declaration of competing interest}
The authors declare that they have no known competing financial interests or personal relationships that could have appeared to influence the work reported in this paper.

\section*{Data availability}
The dataset contains sensitive patient data. We will make synthetic data available upon request.

\section*{Acknowledgements}
This work was supported in part by the National Natural Science Foundation of China (NSFC) under Grant Nos. 62522309, 31871363, and 31900575, and in part by the Natural Science Foundation of Liaoning Province under Grant No. 2025JH6/101000012.

\clearpage
\FloatBarrier
\onecolumn

\appendix

\counterwithin{table}{section}
\renewcommand{\thetable}{\thesection.\arabic{table}}

% Appendix A
\section{Prompts and Output Templates}
\label{app:prompts}

\subsection{System and Instruction Template}
Table~\ref{tab:prompt_template} summarizes the system and instruction template used for CORE-Acu prompting.

\begin{table}[htbp]
  \centering
  \small
  \renewcommand{\arraystretch}{1.25}
  \setlength{\tabcolsep}{6pt}
  \caption{System and Instruction Template.}
  \label{tab:prompt_template}
  \begin{tabularx}{\textwidth}{@{}l X@{}}
    \toprule
    \textbf{Role} & \textbf{Content} \\
    \midrule
    \textbf{System} & You are a TCM acupuncture expert. Follow the required output schema strictly. \\
    \textbf{Instruction} & Based on the patient's symptoms, generate a S-CoT response. Your output must be a single valid JSON object and contain no additional text. \\
    \bottomrule
  \end{tabularx}
\end{table}

\FloatBarrier
\Needspace{34\baselineskip}
\subsection{S-CoT JSON Output Schema}
Table~\ref{tab:json_schema_compact} defines the machine-verifiable JSON output schema (compact view).

\par\smallskip
\noindent\begin{minipage}{\textwidth}
\captionof{table}{JSON Output Schema Specification (Compact View).}
\label{tab:json_schema_compact}
\centering
\footnotesize
\renewcommand{\arraystretch}{1.15}
\setlength{\tabcolsep}{4pt}

  % key columns fixed width; description columns auto-wrap (Y)
  \begin{tabularx}{\textwidth}{@{}p{0.16\textwidth} Y p{0.16\textwidth} Y@{}}
    \toprule
    \multicolumn{2}{c}{\textbf{1. Root Object Fields}} &
    \multicolumn{2}{c}{\textbf{2. Acupoint Object Fields (Nested)}} \\
    \cmidrule(r){1-2} \cmidrule(l){3-4}
    \textbf{Key (Type)} & \textbf{Description} & \textbf{Key (Type)} & \textbf{Description} \\
    \midrule

    \texttt{complaint} (Str)  & Free-text symptom description. &
    \texttt{code} (Str)       & \textbf{Req.} WHO code (e.g., \texttt{GB20}). No duplicates. \\

    \texttt{diagnosis} (List) & TCM syndromes. \textbf{Must be non-empty.} &
    \texttt{name} (Str)       & \textbf{Req.} Pinyin name (e.g., \texttt{Fengchi}). \\

    \texttt{pathology} (Str)  & Pathogenesis reasoning (1--4 sentences). &
    \texttt{side} (Str)       & \textbf{Opt.} Enum: \{\texttt{left}, \texttt{right}, \texttt{bilateral}, \texttt{unspecified}\}. \\

    \texttt{principle} (List) & Treatment principles. \textbf{Must be non-empty.} &
    \texttt{role} (Str)       & \textbf{Opt.} Enum: \{\texttt{main}, \texttt{assist}, \texttt{adjunct}\}. \\

    \texttt{acupoints} (List) & Non-empty list of acupoint objects. &
                             & \textit{(End of nested object)} \\

    \midrule
    \multicolumn{4}{@{}l@{}}{\textbf{3. Global Constraints}} \\
    \multicolumn{4}{@{}p{\textwidth}@{}}{
      \makecell[tl]{%
        \textbullet\ Output must be valid JSON with \textbf{no extra keys} and \textbf{no trailing text}.\\
        \textbullet\ The \texttt{acupoints} list must contain at least one item. Duplicate \texttt{code} entries are prohibited.
      }
    }\\
    \bottomrule
  \end{tabularx}
\end{minipage}
\par\smallskip
\FloatBarrier

\subsection{Concrete Output Example}
Table~\ref{tab:example_instance} provides a concrete S-CoT output example for readability.

\begin{table}[htbpt]
  \centering
  \small
  \renewcommand{\arraystretch}{1.2}
  \setlength{\tabcolsep}{6pt}
  \caption{A concrete S-CoT output example formatted for readability. This example demonstrates how the model explicates the specific \textit{mechanism} for each acupoint selection, linking back to the diagnosis.}
  \label{tab:example_instance}
  \begin{tabularx}{\textwidth}{@{}l X@{}}
    \toprule
    \textbf{Field} & \textbf{Model Output Content} \\
    \midrule
    \textbf{Complaint} & Unilateral temporal (lateral) headache, migraine-like, accompanied by nausea and photophobia. \\
    \addlinespace[4pt]
    \textbf{Diagnosis} & $\bullet$ \textbf{Shaoyang Headache (Migraine):} Unilateral temporal pain linked to the Gallbladder channel. \newline
                         $\bullet$ \textbf{Hyperactivity of Liver-Yang:} Underlying root cause (to be differentiated from Phlegm-Turbidity). \\
    \addlinespace[4pt]
    \textbf{Pathology} & Dysfunction of the \textit{Shaoyang} pivot causing Qi stagnation; Liver Qi rebellion transforming into Fire and disturbing the clear orifices; potentially accompanied by phlegm-turbidity blocking the collaterals. \\
    \addlinespace[4pt]
    \textbf{Principle} & $\bullet$ \textbf{Soothe Shaoyang:} Unblock the pivot mechanism. \newline
                         $\bullet$ \textbf{Calm Liver \& Extinguish Wind:} Address the root cause of rising Yang. \newline
                         $\bullet$ \textbf{Unblock Collaterals:} Relieve local pain. \\
    \addlinespace[4pt]
    \textbf{Acupoints} &
    \begin{itemize}[leftmargin=1.2em, nosep, before=\vspace{-0.5em}, after=\vspace{-0.5em}]
      \item \textbf{SJ5 (Waiguan) + GB41 (Zulinqi):} [Main, Bilateral] \newline
      \textit{Mechanism:} The \textit{Shaoyang} pair. SJ5 dredges the pivot mechanism and clears the head/eyes; GB41 opens the \textit{Dai Mai}. Together they specifically target lateral migraines and reduce sensitivity to external stimuli.
      
      \item \textbf{LR3 (Taichong):} [Main, Bilateral] \newline
      \textit{Mechanism:} \textbf{Targeting the Root.} Courses Liver Qi and pacifies Liver Yang to treat the underlying hyperactivity and reduce the frequency of emotional triggers.
      
      \item \textbf{GB20 (Fengchi):} [Assist, Bilateral] \newline
      \textit{Mechanism:} Dispels wind and unblocks collaterals; pacifies ascending wind-fire to relieve neck tension and head distension.
      
      \item \textbf{GB8 (Shuaigu):} [Assist, Affected Side] \newline
      \textit{Mechanism:} \textbf{Local Point.} Directly enters the temporal collaterals to course wind and stop pain; specifically improves throbbing sensation and photophobia.
    \end{itemize} \\
    \bottomrule
  \end{tabularx}
\end{table}

\FloatBarrier

% Appendix B
\section{Detailed Experimental Results}
\label{app:results}

\subsection{Comprehensive Metric Breakdown}
Table~\ref{tab:merged_metrics} provides a consolidated report of performance. Tables~\ref{tab:negative_feedback_prompt} and \ref{tab:alpaca_example} detail the rewriting template and SFT data record.

\begin{table*}[!t]
  \centering
  \footnotesize
  \renewcommand{\arraystretch}{1.1}
  \setlength{\tabcolsep}{0pt}
  \caption{Comprehensive breakdown of automated metrics on the held-out test set.}
  \label{tab:merged_metrics}
  \begin{tabular*}{\textwidth}{@{\extracolsep{\fill}}l cccc ccc ccc @{}}
    \toprule
    \multirow{2}{*}{\textbf{Model}} & \multicolumn{4}{c}{\textbf{BLEU Scores}} & \multicolumn{3}{c}{\textbf{ROUGE Scores}} & \multicolumn{3}{c}{\textbf{Entity Metrics}} \\
    \cmidrule(lr){2-5} \cmidrule(lr){6-8} \cmidrule(lr){9-11}
     & B-1 & B-2 & B-3 & B-4 & R-1 & R-2 & R-L & Prec. & Rec. & F1 \\
    \midrule
    GPT-4o & 0.3470 & 0.1991 & 0.1205 & 0.0722 & 0.2753 & 0.0961 & 0.1985 & 0.3812 & 0.2368 & 0.2906 \\
    Qwen-Max & 0.3788 & 0.2259 & 0.1401 & 0.0865 & 0.2949 & 0.1054 & 0.1988 & 0.3678 & 0.2679 & 0.3087 \\
    Qwen3-8B & 0.3735 & 0.2159 & 0.1343 & 0.0814 & 0.2865 & 0.1025 & 0.1883 & 0.3182 & 0.2970 & 0.3062 \\
    GLM-4.6 & 0.3655 & 0.2061 & 0.1193 & 0.0697 & 0.3257 & 0.1137 & 0.1970 & 0.3480 & 0.3702 & 0.3511 \\
    HuatuoGPT  & 0.0331 & 0.0188 & 0.0107 & 0.0059 & 0.2054 & 0.0443 & 0.1209 & 0.4569 & 0.1431 & 0.2161 \\
    \textbf{CORE-Acu (Ours)} & \textbf{0.4661} & \textbf{0.3330} & \textbf{0.2471} & \textbf{0.1898} & \textbf{0.4522} & \textbf{0.2472} & \textbf{0.3334} & \textbf{0.5576} & \textbf{0.3955} & \textbf{0.4612} \\
    \bottomrule
  \end{tabular*}
  \begin{tabular*}{\textwidth}{@{}l@{}}
    \scriptsize \textit{Note: \textbf{B-1-4}: BLEU-1 to BLEU-4; \textbf{R-1/2/L}: ROUGE-1, ROUGE-2, and ROUGE-L; \textbf{Prec.}: Precision; \textbf{Rec.}: Recall; \textbf{F1}: Entity-F1 Score.}
  \end{tabular*}
\end{table*}

\begin{table*}[!t]
  \centering
  \footnotesize
  \caption{Template Specification for Violation-Aware Rewriting.}
  \label{tab:negative_feedback_prompt}
  \renewcommand{\arraystretch}{1.1}
  \begin{tabularx}{\textwidth}{@{}l X@{}}
    \toprule
    \textbf{Component} & \textbf{Content / Specification} \\
    \midrule
    \textbf{Trigger Context} & Given generated acupoints $\mathcal{A}$ and a detected conflicting tuple $(x,r,y)$ where $r \in \mathcal{R}_{\text{unsafe}}$. \\
    \addlinespace[2pt]
    \textbf{System Alert} & \textbf{Safety Verifier Alert:} Detected unsafe violation under \texttt{<RULE\_TYPE>}. \\
    \addlinespace[2pt]
    \textbf{Evidence Injection} & \textbf{Conflict evidence:} \texttt{<EVIDENCE>} (\textit{e.g., \{SJ5, GB41\} is Incompatible}) \\
    \addlinespace[2pt]
    \textbf{Task Instruction} & \textbf{Your task:} Revise \texttt{acupoints} to eliminate conflict while keeping others unchanged. \\
    \addlinespace[2pt]
    \textbf{Hard Constraints} & 
    \begin{minipage}[t]{\linewidth}
      \begin{enumerate}[label=(\arabic*), leftmargin=1.2em, nosep, before=\vspace{0pt}]
        \item Output \textbf{only} a valid JSON object; (2) No text outside JSON; (3) No conflicting pairs.
      \end{enumerate}
    \end{minipage} \\
    \addlinespace[2pt]
    \textbf{Fallback Policy} & If unresolved within $T_{\max}$ retries, terminate and request human confirmation. \\
    \bottomrule
  \end{tabularx}
\end{table*}

\begin{table*}[!t]
  \centering
  \footnotesize
  \caption{Example of a Training Instance in Alpaca Format (SFT Data Record).}
  \label{tab:alpaca_example}
  \renewcommand{\arraystretch}{1.05}
  \begin{tabularx}{\textwidth}{@{}l X@{}}
    \toprule
    \textbf{Field} & \textbf{Content} \\
    \midrule
    \textbf{Instruction} & Perform syndrome differentiation using TCM acupuncture theory and provide structured S-CoT. \\
    \addlinespace[2pt]
    \textbf{Input} & Unilateral temporal (lateral) headache, migraine-like \\
    \addlinespace[2pt]
    \textbf{Output} &
    \begin{minipage}[t]{\linewidth}
      \textbf{\textit{Target S-CoT Response:}}
      \begin{description}[leftmargin=0pt, itemsep=0pt, parsep=1pt, font=\bfseries]
        \item[[complaint]:] Unilateral temporal (lateral) headache, migraine-like
        \item[[diagnosis]:] Shaoyang (lateral/temporal) headache / Migraine; Hyperactivity of liver-yang (to be differentiated).
        \item[[pathology]:] Dysfunction of the Shaoyang pivot, stagnation of liver and gallbladder Qi transforming into fire.
        \item[[principle]:] Soothe Shaoyang; Calm the liver and suppress yang; Unblock collaterals to relieve pain.
        \item[[acupoints]:] \textbf{SJ5, GB41}: \textit{main}; \textbf{GB20, GB8}: \textit{assist}.
      \end{description}
    \end{minipage} \\
    \bottomrule
  \end{tabularx}
\end{table*}

\FloatBarrier

% Appendix C
\section{LLM-as-a-Judge Evaluation Protocol}
\label{app:judge}

\subsection{Scoring Rubric and Prompt Structure}
\label{sec:judge_rubric}
Table~\ref{tab:scoring_rubric} defines the five clinical evaluation dimensions (0--10 each).

\begin{table}[htbpt]
  \centering
  \small
  \renewcommand{\arraystretch}{1.20}
  \setlength{\tabcolsep}{6pt}
  \caption{Definition of the five clinical evaluation dimensions (Scale: 0--10 per dimension).}
  \label{tab:scoring_rubric}
  \begin{tabularx}{\textwidth}{@{} >{\bfseries}l X @{}}
    \toprule
    \textbf{Dimension} & \textbf{Assessment Criteria} \\
    \midrule
    Reasoning Validity & Assess if the reasoning chain is coherent and free of internal contradictions. \\
    Dialectical Accuracy & Evaluate whether the TCM syndrome (diagnosis) is correct and reflects the precise nature/location of the condition. \\
    Pathological Logic & Verify if the pathological mechanism explanation aligns with TCM theory. \\
    Prescription Rationality & Assess whether selected acupoints align with therapeutic principles and include critical main points. \\
    Explicability & Evaluate transparency, structural clarity, and professional standardization of the output. \\
    \bottomrule
  \end{tabularx}
\end{table}

Table~\ref{tab:judge_prompt_structure} summarizes the judge prompt structure used for evaluation.

\begin{table}[htbpt]
  \centering
  \small
  \renewcommand{\arraystretch}{1.20}
  \setlength{\tabcolsep}{6pt}
  \caption{Structure of the Judge Prompt used for evaluation.}
  \label{tab:judge_prompt_structure}
  \begin{tabularx}{\textwidth}{@{}l X@{}}
    \toprule
    \textbf{Component} & \textbf{Content / Specification} \\
    \midrule
    \textbf{System Role} & You are a senior TCM acupuncture expert and professor. You need to compare the answers of multiple candidates and score each candidate based on five specific dimensions. \\
    \textbf{Scoring Criteria} & \textit{(Detailed rubric as defined in Section~\ref{sec:judge_rubric})} \\
    \textbf{Input Data} & The prompt receives a formatted block of answers: \newline
    \texttt{=== Candidate Answer List ===} \newline
    \texttt{\{candidates\_text\_block\}} \\
    \textbf{Output Req.} & 1) Compare all candidates to ensure scoring consistency; \newline
    2) Output JSON only (no Markdown); \newline
    3) Keys must match the input names exactly. \\
    \bottomrule
  \end{tabularx}
\end{table}

\subsection{Sample Judge Output}
Table~\ref{tab:judge_sample} presents a sample structured output returned by the judge LLM.

\begin{table*}[!t]
  \centering
  \caption{Detailed Evaluation Output from Judge LLM (Sample Case 001). Scores represent five key TCM dimensions (0--10).}
  \label{tab:judge_sample}
  \small
  \renewcommand{\arraystretch}{1.0} % 收紧纵向空间
  \setlength{\tabcolsep}{5pt}
  \begin{tabularx}{\textwidth}{l ccccc c X}
    \toprule
    \textbf{Model} & \textbf{Reas.} & \textbf{Dial.} & \textbf{Path.} & \textbf{Pres.} & \textbf{Expl.} & \textbf{Total} & \textbf{Judge's Rationale (Translated)} \\
    \midrule
    \textbf{CORE-Acu} & \textbf{10} & \textbf{9} & \textbf{10} & \textbf{9} & \textbf{10} & \textbf{48} & The reasoning chain is highly coherent, accurately capturing the root-branch relationship. The combination of main points is rigorous and principles-based. \\
    GLM-4.6 & 9 & 9 & 9 & 9 & 8 & 44 & Demonstrates a complete chain from etiology to treatment. Syndrome differentiation is precise, though expanded therapies are slightly generalized. \\
    GPT-4o & 8 & 8 & 8 & 8 & 8 & 40 & The differentiation of phlegm-qi binding is accurate with smooth logic. The prescription aligns well, though additions appear somewhat formulaic. \\
    Qwen-Max & 8 & 7 & 7 & 7 & 8 & 37 & Overall structure is clear, but syndrome boundaries are slightly overlapping. The point significance explanation lacks depth for certain key acupoints. \\
    HuatuoGPT & 6 & 5 & 5 & 5 & 6 & 27 & Syndrome differentiation is overly coarse. The prescription is brief with insufficient justification for specific point selection. \\
    Qwen3-8B & 4 & 4 & 4 & 3 & 5 & 20 & Internal logic is loose. Multiple significant errors in acupoint meridian attributes and anatomical locations were detected. \\
    \bottomrule
    \multicolumn{8}{l}{\scriptsize \textit{Note: Reas: Reasoning; Dial: Dialectical; Path: Pathology; Pres: Prescription; Expl: Explicability.}}
  \end{tabularx}
\end{table*}

\FloatBarrier

\begin{table}[htbp]
    \centering
    \normalsize 
    \renewcommand{\arraystretch}{1.3}
    
    \caption{Data Schema of the Proposed Dataset}
    \label{tab:schema}
    
    \begin{tabularx}{\linewidth}{l l X}
        \toprule
        \textbf{Field Key} & \textbf{Data Type} & \textbf{Description} \\
        \midrule
        
        complaint & String & Free-text description of patient symptoms. \\
        
        diagnosis & List[String] & TCM syndrome differentiation (Bian-Zheng). \\
        
        pathology & String & Mechanism explanation of the illness. \\
        
        principle & List[String] & Treatment principles guiding the prescription. \\
        
        acupoints & List[Object] & Structured acupoint prescription. Each object contains: \newline
         $\bullet$ code: acupoint ID\newline
         $\bullet$ name: Pinyin name\newline
         $\bullet$ side: Specific side\newline
         $\bullet$ role: Therapeutic role \\
        \bottomrule
    \end{tabularx}
\end{table}

\twocolumn
\bibliography{sample}

\end{document}